\documentclass[preprint,12pt]{elsarticle}

\usepackage{amssymb}
\usepackage{amsmath}
 \usepackage{verbatim} 
\usepackage{algorithm}
\usepackage{algpseudocode}
\usepackage{amsmath}
\usepackage[utf8]{vietnam} 
\usepackage{longtable}
\usepackage{hyperref}
\usepackage[english]{babel}
\usepackage{float}
\usepackage{lineno}
\usepackage{makecell}

\journal{Journal of King Saud University Computer and Information Sciences}

\begin{document}
\sloppy
\begin{frontmatter}



\title{Explainable Multi-Loss Distillation Framework for Efficient and Interpretable Shrimp Disease Text Classification} 

 \author[a]{Anh Nguyen Quynh\fnref{equal}}
 \affiliation[a]{organization={Research and Development Application Department, FPT University},
             addressline={},
             city={Cantho city},
             postcode={90000},
             state={},
             country={Vietnam}}
 \author[b]{Khang Nguyen Quoc\fnref{equal}}
 \affiliation[b]{organization={School of Electrical Engineering, Korea University},
             addressline={},
             city={Seoul city},
             postcode={02841},
             state={},
             country={South Korea}}
\fntext[equal]{These authors contributed equally to this work.}

\author[c]{Luyl-Da Quach\corref{cor1}}
\cortext[cor1]{Corresponding author: luyldaquach@gmail.com}
 \affiliation[c]{organization={Department of Information Technology, FPT University},
             addressline={},
             city={Cantho city},
             postcode={90000},
             state={},
             country={Vietnam}}

\begin{abstract}
Shrimp disease classification has become an urgent issue due to its significant impact on the import–export output of producing countries, particularly Vietnam. Most existing studies focus on image-based classification, which typically operates at the late stage of disease manifestation. Therefore, text-based classification has the potential to enable early and timely disease detection. To address this limitation, we introduce the SALT (\textbf{S}hrimp disease text \textbf{A}nalysis with multi-\textbf{L}oss dis\textbf{T}illation) framework, which incorporates explainability analysis using Local Interpretable Model-agnostic Explanations (LIME) and SHapley Additive exPlanations (SHAP) to evaluate model predictions and interpret the learned linguistic features. Experimental results demonstrate that SALT achieves competitive performance across multiple distillation objectives, outperforming supervised baselines while providing a favorable trade-off between predictive performance and computational efficiency. Moreover, it exhibits strong explainability, accurately identifying key linguistic features and semantic patterns relevant to disease descriptions. These findings highlight the potential of knowledge distillation–based text classification for future applications in early shrimp disease diagnosis and related research directions. 
\end{abstract}

\begin{graphicalabstract}
\includegraphics[width=1\linewidth]{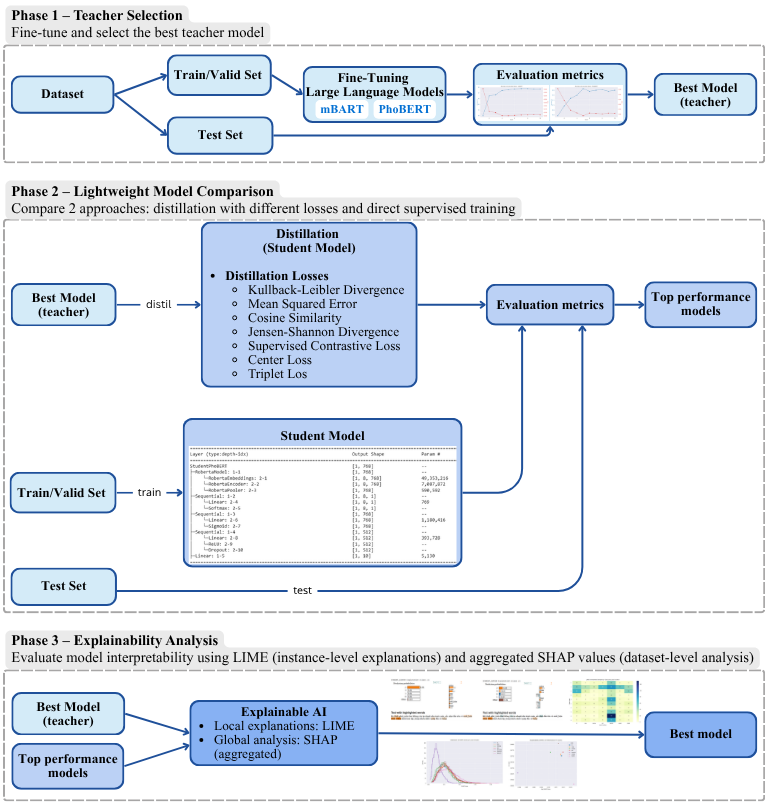}
\end{graphicalabstract}

\begin{highlights}
\item Introduction of SALT, an explainable distillation framework that unifies teacher-student learning with explainability-driven evaluation for Vietnamese domain-specific text understanding, while systematically investigating multiple distillation objectives through a comprehensive comparison of distribution-based and metric-learning losses
\item Integration of local explanations using LIME, global feature attribution analysis using aggregated SHAP values, and a keyword deletion test to evaluate explanation faithfulness.
\item Comprehensive experimental evaluation comparing multiple distillation objectives and strong baselines in terms of classification performance, computational efficiency, and explanation analysis.

\end{highlights}


\end{frontmatter}



\section{Introduction}

According to the May 2025 report by the Food and Agriculture Organization (FAO) of the United Nations, the global shrimp production exceeded 3 million tons, with Vietnam ranked among the top three shrimp-exporting countries worldwide~\cite{Ref15}. However, this achievement is accompanied by significant challenges related to disease outbreaks in shrimp farming, which severely impact productivity. 

Recognizing this issue, numerous studies have employed AI to diagnose shrimp diseases through image analysis using Convolutional Neural Networks (CNNs) and Transfer Learning (TL) techniques. \citet{Ref16} implemented InceptionV3 and MobileNet models to classify images of diseased shrimp, achieving an accuracy of approximately 90\%. \citet{Ref17} compared VGG16, EfficientNetB0, and ResNet50 architectures for identifying White Spot Syndrome Virus and reported that ResNet50 achieved the highest accuracy of 90.43\%. \citet{Ref18} also proposed an improved LeNet model for rapid detection of shrimp diseases. In addition, several studies have explored text-based diagnosis of shrimp disease, in which symptom descriptions are analyzed using natural language processing (NLP) techniques, such as the Term Frequency-Inverse Document Frequency (TF–IDF) method, for disease classification ~\cite{Ref19}. Finally, the text-based study demonstrated the image-text model's significant contribution to improving video Question and Answer results \cite{huang2024global}. These approaches demonstrate that deep learning (DL) algorithms have also yielded promising results in this field. However, existing studies still primarily focus on model accuracy, with limitations including insufficient attention to model compactness, interpretability, and real-world environments.

In recent years, pre-trained language models (PLMs) have advanced significantly and become the dominant paradigm for text classification tasks~\citep{Ref1, Ref02}. In particular, Bidirectional Encoder Representations from Transformers (BERT) and its Vietnamese counterpart PhoBERT~\citep{Ref03} have demonstrated strong performance by effectively capturing bidirectional contextual semantics. Although these models have achieved success across a wide range of Vietnamese NLP tasks, their application to domain-specific agricultural and aquaculture problems remains limited, particularly for symptom-based shrimp disease diagnosis from Vietnamese textual descriptions, where both domain-specific semantic understanding and computational efficiency are essential for practical deployment. Nevertheless, PLMs are often computationally expensive, especially when trained on large-scale datasets. Moreover, these models often operate as “black boxes” offering limited insight into their internal decision-making processes and feature attributions. In real-world applications, textual symptom descriptions are inherently noisy and domain-dependent, which can compromise model reliability, yield untrustworthy predictions, and cause significant performance degradation, particularly in high-stakes domains such as agriculture, healthcare, and aquaculture. Consequently, enhancing the interpretability and robustness of language models has become a crucial requirement for deploying domain-specific, trustworthy NLP systems in real-world contexts.

Knowledge distillation (KD) has recently emerged as an effective model compression technique that transfers the knowledge of large-scale teacher models into more compact student models, offering a practical pathway to leverage the representational power of large language models in resource-constrained settings. Within the NLP domain, this paradigm has been widely explored across diverse architectures and objectives. For instance, a multi-teacher framework incorporating BERT, BiGRU, and TextCNN has been shown to enhance classification accuracy while substantially reducing computational overhead~\citep{Ref04}, whereas another study demonstrated that KD with BERT can effectively mitigate catastrophic forgetting, yielding superior downstream performance~\citep{Ref05}. Building upon these efforts, \citet{Ref06} proposed a BERT-based pre-training framework combined with an adaptive distillation strategy to enhance the student model's learning efficiency. However, most existing KD approaches primarily focus on improving model performance, with limited emphasis on interpretability and generalization~\cite{Ref07, Ref08, Ref09}. Several studies have focused on expanding business models using various approaches across different fields, such as \citet{Ref20} focus on preserving structural knowledge and feature relationships, \citet{Ref21} employs contrastive representation learning to enhance generalization capability, \citet{Ref22, Ref23} has been widely adopted to compress large transformer models into smaller architectures such as DistilBERT and TinyBERT, \citet{Ref24, Ref25} have applied KD to optimize transformer models in NLP, such as Patient-KD and MiniLM, which focus on improving the efficiency of knowledge transfer in the self-attention layers of the transformer, thereby helping to optimize the learning process of the student model. Despite promising results, most KD work in NLP still focuses on model reduction without considering how to ensure transparency and interpretability in the knowledge transfer process. Therefore, combining KD with XAI is a promising approach to developing lightweight, transparent models suitable for sensitive problems, such as disease diagnosis in aquaculture, especially in classification tasks using NLP.

Explainable Artificial Intelligence (XAI) has become an important research area focused on improving the interpretability of DL models, enabling humans to better understand the internal decision-making mechanisms of these systems. Several studies applied XAI to agricultural image data, such as maize leaf disease classification~\cite{Ref10}, rice pest detection~\cite{Ref11}, demonstrating promising results in elucidating disease-related visual features. In addition, various research efforts in the NLP domain have achieved encouraging results by leveraging XAI to automatically analyze and interpret textual features for classification and interpretability tasks. For instance, sentiment classification and local explanation in education~\cite{Ref12}, feature-level word attribution for sarcasm detection~\cite{Ref13}, and identifying hate and aggression-related expressions in social media~\citep{Ref14}. In NLP, \citet{Ref28} built the ERASER benchmark to quantify the degree of connection between the explanation and the prediction mechanism of the model, \citet{Ref29} introduced an evaluation framework for faithfulness in NLP systems, \citet{Ref30} applied SHAP to explain the decisions of autoregressive models in text generation tasks, showing the potential of XAI in large language models. From these studies, XAI techniques are widely applied in NLP tasks such as Local Interpretable Model-Agnostic Explanations (LIME) \cite{Ref26} and SHapley Additive exPlanations (SHAP) \cite{Ref27}, including text classification, natural language inference, and question-answering, to assess model interpretability. Collectively, these studies highlight the crucial role of XAI in enhancing the transparency and reliability of model predictions. Despite these advances, integrating XAI into the analysis and interpretive evaluation of KD limitations remains a challenge.

Overall, the challenges in NLP-based text classification center on computational efficiency, the limited availability of domain-specific Vietnamese language resources, and the limited interpretability of black-box models. Therefore, we introduce SALT (Shrimp disease text Analysis with multi-Loss disTillation), a teacher–student learning framework that integrates explainability analysis using LIME and SHAP into knowledge distillation to facilitate the interpretation of model predictions. Our main contributions are summarized as follows:

\begin{itemize}
    \item Introduction of SALT, an explainable distillation framework that unifies teacher-student learning with explainability-driven evaluation for Vietnamese domain-specific text understanding, while systematically investigating multiple distillation objectives through a comprehensive comparison of distribution-based and metric-learning losses
    \item Integration of local explanations using LIME, global feature attribution analysis using aggregated SHAP values, and a keyword deletion test to evaluate explanation faithfulness.
    \item Comprehensive experimental evaluation comparing multiple distillation objectives and strong baselines in terms of classification performance, computational efficiency, and explanation analysis.
\end{itemize}

\section{Materials and method}
\subsection{Data collection and preprocessing}\label{dataset}
In this study, we employ the ShrimpCap dataset, which comprises $1,000$ Vietnamese symptom descriptions covering $10$ common shrimp diseases: Acute Hepatopancreatic Necrosis Disease, White Feces Disease, White Spot Disease, Luminous Bacterial Disease, Loose Shell Syndrome, Black Gill Disease, Filamentous Bacterial Disease, Vitamin C Deficiency Disease, Yellow Head Disease, and Taura Syndrome. ShrimpCap dataset was curated from publicly available Vietnamese aquaculture resources, including disease management guidelines and online forums discussing shrimp disease symptoms and treatments. The collected descriptions were manually assigned to disease categories and reviewed by expert collaboration in aquaculture to verify the accuracy of symptom-label alignment. Our dataset is evenly distributed across 10 disease classes, providing semantically diverse descriptions, as summarized in Table~\ref{tab:vn-en-symptoms}.

\begin{longtable}{|p{0.05\linewidth}|p{0.35\linewidth}|p{0.45\linewidth}|}
\caption{Vietnamese symptom descriptions and their English translations.}
\label{tab:vn-en-symptoms} \\
\hline
\textbf{\#} & \textbf{Vietnamese} & \textbf{English} \\
\hline
\endfirsthead

\hline
\textbf{\#} & \textbf{Vietnamese} & \textbf{English} \\
\hline
\endhead

\hline
\endfoot

1 & Khi bệnh tiến triển, gan tụy từ màu vàng nhạt sang trắng đục, ban đầu sưng nhưng teo dần theo thời gian. Trên bề mặt có nhiều đốm đen, ruột tôm trống không, cơ thể suy yếu và chết hàng loạt. & As the disease progresses, the hepatopancreas changes from pale yellow to opaque white, initially swollen but gradually shrinking over time. Numerous black spots appear on the surface, the intestine becomes empty, and the shrimp's body weakens, leading to mass mortality. \\

2 & Tôm bị bệnh có ruột yếu, thức ăn không hấp thụ hết, phân chứa nhiều chất nhầy, nổi trên bề mặt ao. & Diseased shrimp have fragile intestines, incomplete feed absorption, and mucus-laden feces that float on the pond surface. \\

3 & Tôm bơi yếu, tập trung gần mép bờ và có dấu hiệu cơ thịt hơi đục khi quan sát. & Shrimp exhibit weak swimming behavior, gather near the pond edges, and display slightly opaque musculature upon observation. \\

4 & Quan sát thấy tôm có rong bám ở mang, vỏ nhớt, tôm chết rải rác dưới đáy ao, số lượng tăng dần theo mức độ bệnh. & 
Shrimp are observed with algal attachments on the gills, a mucous carapace, and scattered mortalities on the pond bottom, which increase in frequency as the disease progresses. \\

5 & Tôm bơi không linh hoạt, thường tụ lại ở các góc ao hoặc vùng nước cạn, mang có dấu hiệu phì đại, kèm theo màu nâu hoặc đen dần rõ rệt. & Shrimp swim sluggishly, often cluster in pond corners or shallow areas. The gills appear hypertrophied, with a gradual darkening to brown or black coloration. \\

6 & Quan sát thấy tôm khó lột xác, chân bơi bị bao phủ bởi lông tơ, mang có màu nâu hoặc đen. & Molting becomes difficult; pleopods are covered with fine filaments, and the gills turn brown or black. \\

7 & Quan sát thấy râu và chân của tôm không còn màu sáng tự nhiên, thay vào đó là các mảng đen loang lổ. & The antennae and pereiopods lose their natural bright color, exhibiting irregular black patches instead. \\

8 & Khi mắc bệnh, tôm có dấu hiệu di chuyển mất phương hướng, tập trung tại vùng nước ít lưu thông, thân nhợt nhạt hơn bình thường. & When infected, shrimp exhibit disoriented movement, congregate in areas with poor water circulation, and display paler body coloration than normal. \\

9 & Chân bò tôm chuyển đỏ, vỏ trở nên mềm yếu, ruột không có thức ăn, tôm phát triển chậm và dễ chết. & The pereiopods become reddish, the carapace softens, the gut is empty, growth slows, and mortality risk increases. \\

10 & Ruột rỗng, tôm nhạt màu, yếu. Gan tụy lúc đầu sưng nhẹ, sau đó nhỏ lại, có đốm đen. Tôm bỏ ăn, chết nhiều khi bệnh nặng. & The gut is empty, shrimp appear pale and weak. The hepatopancreas is slightly swollen at first, then shrinks and develops black spots. Shrimp stop feeding and experience high mortality in severe disease stages. \\
\hline
\end{longtable}

\begin{figure}[H]
    \centering
    \includegraphics[width=1\linewidth]{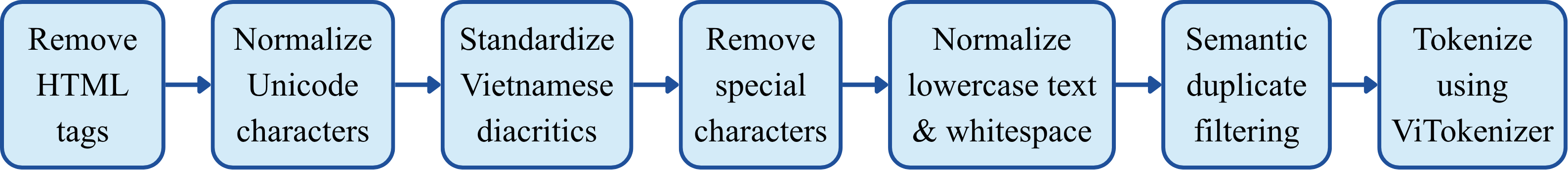}
    \caption{Data preprocessing pipeline for disease descriptions.}
    \label{fig1}
\end{figure}

\begin{figure}[H]
    \centering
    \includegraphics[width=1\linewidth]{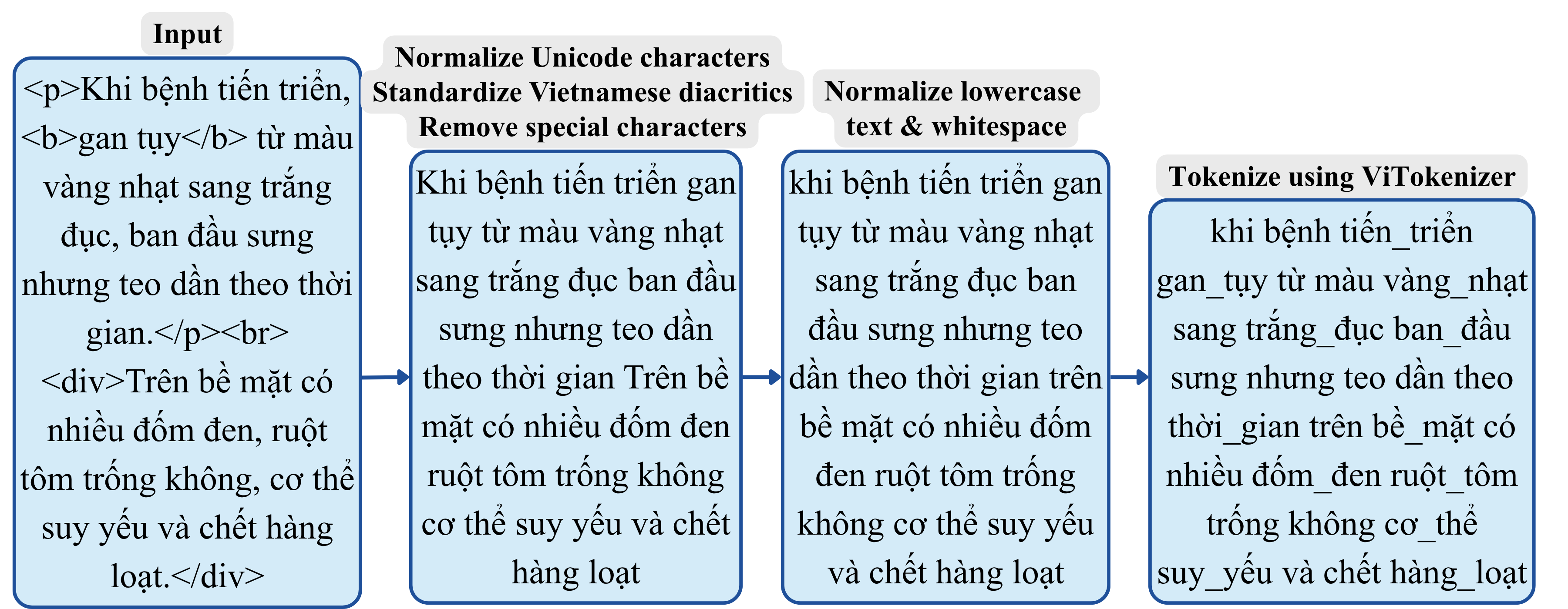}
    \caption{Text transformation of a specific description under our pipeline.}
    \label{fig2}
\end{figure}

For data preprocessing, we applied a comprehensive pipeline to standardize input data, remove duplicate and semantically redundant descriptions, and ensure data consistency to minimize bias during model training (Figure~\ref{fig1}).To ensure vocabulary consistency and prevent spurious features, we define the pipeline including: (1) data cleaning by removing HTML tags, (2) normalizing Unicode characters, and (3) standardizing Vietnamese diacritics. Moreover, special characters without semantic meaning are removed to reduce noise. These texts are converted to lowercase, and excess whitespace is removed to ensure consistent formatting. Then, semantic duplicate filtering is performed using sentence embeddings. Additionally, we compute the similarity between these descriptions and remove similar descriptions by a threshold of 0.95. After this step, we also manually inspected and either removed or revised to reduce semantic redundancy. Lastly, word segmentation is performed using ViTokenizer~\citep{Ref31} as Vietnamese contains many meaningful compound words. An illustrative example of how a raw description is transformed by the pipeline is shown in Figure~\ref{fig2}.

\begin{figure} [H]
    \centering
    \includegraphics[width=0.75\linewidth]{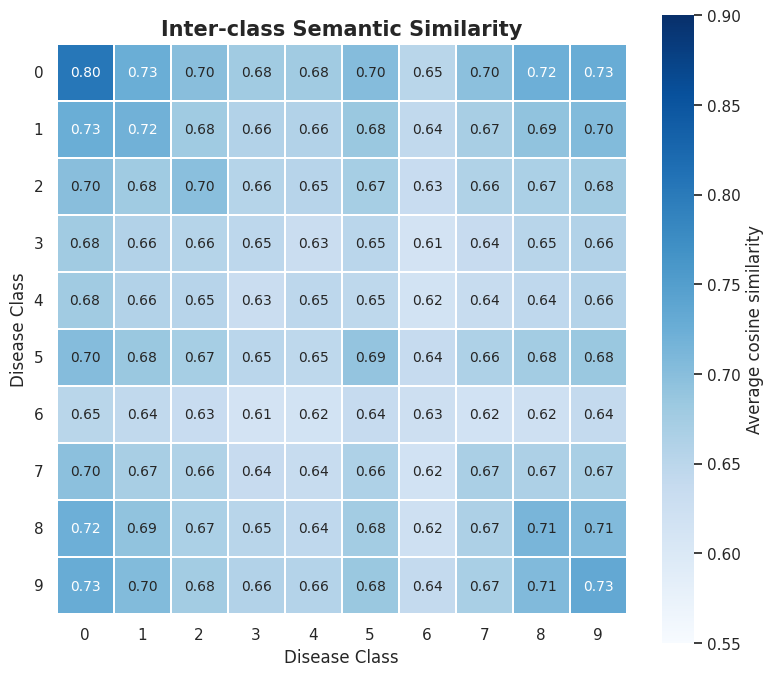}
    \caption{Inter-class semantic similarity matrix based on the average pairwise cosine similarity of sentence embeddings.}
    \label{fig3}
\end{figure}

For inter-class semantic analysis, we computed a similarity matrix from sentence embeddings of the disease descriptions. As shown in Figure~\ref{fig3}, we observed high semantic similarity across several disease pairs, indicating that multiple shrimp diseases share common symptom descriptions.

After this processing, we present the ShrimpCap dataset, comprising $979$ symptom descriptions, which was partitioned into training, validation, and test sets following a $6:2:2$ ratio for our experiment, as summarized in Table~\ref{tab:data-statistics}.

\begin{table}[H]
\centering
\caption{Data statistics of the dataset after preprosessing.}
\label{tab:data-statistics}
\resizebox{\linewidth}{!}{%
\begin{tabular}{lcccc}
\hline
\textbf{Class} & \textbf{Train} & \textbf{Valid} & \textbf{Test} & \textbf{Total} \\
\hline
Acute Hepatopancreatic Necrosis Disease & 56 & 19 & 19 & 94 \\
White Feces Disease & 60 & 20 & 20 & 100 \\
White Spot Disease & 60 & 20 & 20 & 100 \\
Luminous Bacteria Disease & 58 & 19 & 19 & 96 \\
Loose Shell Syndrome & 59 & 19 & 20 & 98 \\
Black Gill Disease & 60 & 20 & 20 & 100 \\
Filamentous Bacterial Disease & 59 & 20 & 20 & 99 \\
Vitamin C Deficiency Disease & 55 & 19 & 18 & 92 \\
Yellow Head Disease & 60 & 20 & 20 & 100 \\
Taura Syndrome & 60 & 20 & 20 & 100 \\
\hline
\textbf{Total} & \textbf{587} & \textbf{196} & \textbf{196} & \textbf{979} \\
\hline
\end{tabular}%
}
\end{table}

\vspace{0.5em}
\subsection{Augmentation for baseline evaluation.}
\begin{figure}[H]
    \centering
    \includegraphics[width=0.78\linewidth]{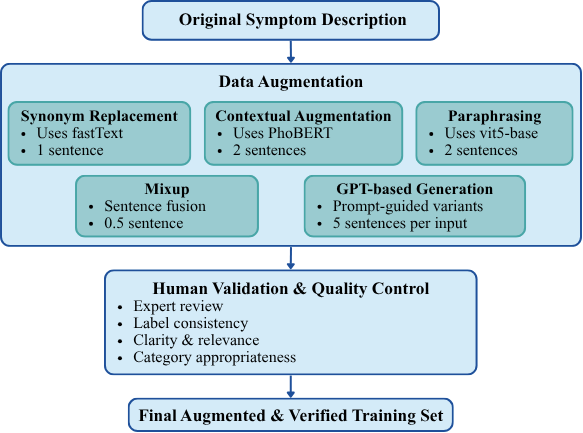}
    \caption{Workflow for constructing the augmented training set used in the linguistic augmentation baseline.}
    \label{fig4}
\end{figure}

To improve ShrimpCap dataset, we propose an augmented data on the training set by constructing exclusively for the linguistic augmentation baseline. This augmented data were generated to investigate whether conventional linguistic augmentation could further improve the student model's performance. Hence, we applied four method-based augmentation strategies: 
\begin{itemize}
    \item \textbf{Synonym Replacement.} We randomly replace these words with their synonyms, based on our constructed dictionary from the dataset and manually edited prior to replacement, ensuring accuracy and context suitability.
    \item \textbf{Contextual Augmentation.} Generating semantically equivalent sentences by selectively replacing words with contextually appropriate alternatives predicted by PhoBERT, thereby preserving the original meaning while introducing lexical diversity.
    \item \textbf{Paraphrasing.} With VietAI/vit5\footnote{https://huggingface.co/VietAI/vit5-base}base model, creating many semantically expressions for each original caption, thereby enriching the grammatical structure and increasing the diversity of the data (Figure~\ref{fig4}).
    \item \textbf{Symptom Mix-up.} We apply mix-up technique by combining two symptom descriptions from the same disease class to create synthetic sentences, thereby helping the model learn the general characteristics of each disease group.
\end{itemize}

\begin{figure}[H]
    \centering
    \includegraphics[width=0.8\linewidth]{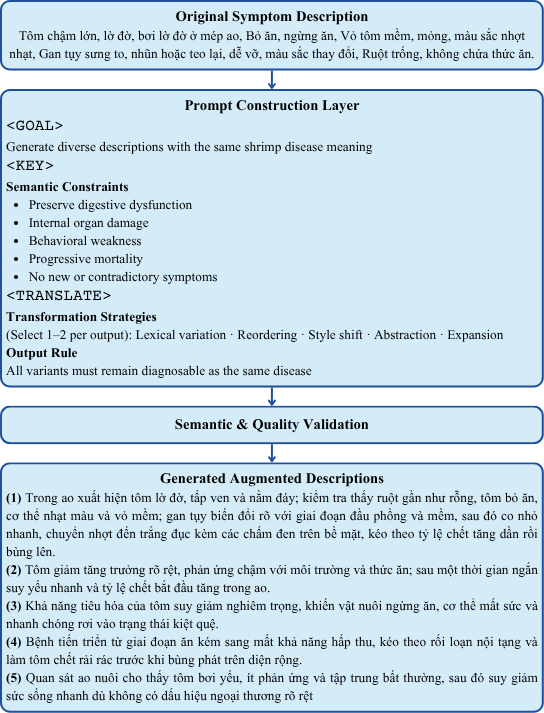}
    \caption{Prompt-guided linguistic augmentation pipeline for generating additional shrimp disease descriptions.}
    \label{fig5}
\end{figure}

Moreover, we designed a prompt-guided augmentation framework based on GPT-5.5~\citep{gpt5} generate new descriptions with diverse wording and writing styles (Figure~\ref{fig5}). Our prompt consists of three components: \texttt{<GOAL>} component specifies the overall objective of generating semantically equivalent symptom descriptions, \texttt{<KEY>} component preserves essential disease characteristics while preventing contradictory or fabricated symptoms, and \texttt{<TRANSLATE>} component controls stylistic transformations, including lexical substitution, sentence reordering, style shifting, abstraction, and expansion, while ensuring that every generated description remains diagnostically consistent with the original disease. 
These generated descriptions were manually reviewed with expert collaboration before being incorporated into the augmented training set. The expert verified semantic consistency, preservation of disease labels, linguistic clarity, and absence of symptoms. These descriptions that failed to satisfy these criteria were revised or discarded before inclusion. After augmentation, the expanded training set was combined with the unchanged validation and test sets to form the augmented ShrimpCap dataset used exclusively for the linguistic augmentation baseline. The resulting data distribution is summarized in Table~\ref{tab:augmented-data-statistics}.

\begin{table}[H]
\centering
\caption{Data statistics of the ShrimpCap dataset with an augmented training set.}
\label{tab:augmented-data-statistics}
\resizebox{\linewidth}{!}{%
\begin{tabular}{lcccc}
\hline
\textbf{Class} & \textbf{Train} & \textbf{Valid} & \textbf{Test} & \textbf{Total} \\
\hline
Acute Hepatopancreatic Necrosis Disease & 332 & 19 & 19 & 370 \\
White Feces Disease & 360 & 20 & 20 & 400 \\
White Spot Disease & 350 & 20 & 20 & 390 \\
Luminous Bacteria Disease & 344 & 19 & 19 & 382 \\
Loose Shell Syndrome & 354 & 19 & 20 & 393 \\
Black Gill Disease & 359 & 20 & 20 & 399 \\
Filamentous Bacterial Disease & 353 & 20 & 20 & 393 \\
Vitamin C Deficiency Disease & 329 & 19 & 18 & 366 \\
Yellow Head Disease & 360 & 20 & 20 & 400 \\
Taura Syndrome & 357 & 20 & 20 & 397 \\
\hline
\textbf{Total} & \textbf{3,498} & \textbf{196} & \textbf{196} & \textbf{3,890} \\
\hline
\end{tabular}%
}
\end{table}

\subsection{Method}
\begin{figure}[H]
    \centering
    \includegraphics[width=1\linewidth]{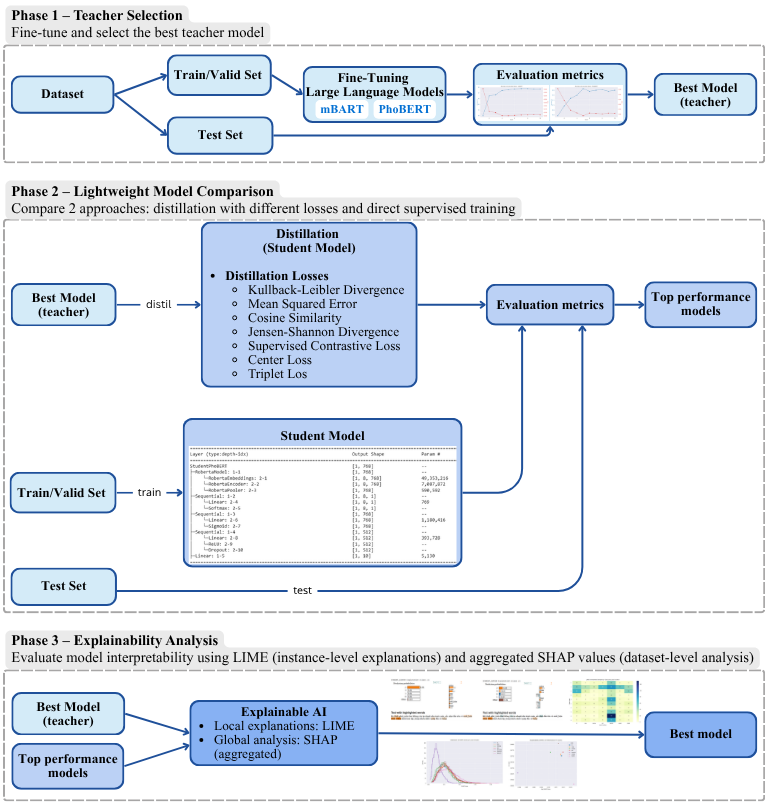}
    \caption{Overall framework of model distillation, comparison, and explainability.}
    \label{fig6}
\end{figure}

In this study, we propose an overall framework of the teacher–student learning model based on the distillation illustrated in Figure~\ref{fig6}. Our framework includes three main phases: 

\begin{itemize}
    \item \textbf{Phase 1: Teacher Selection.} Our objective is to identify the most suitable teacher model. State-of-the-art models are fine-tuned on the training set, with the validation set used for model selection and training monitoring. These trained models are then evaluated on the test set using evaluation metrics to select the best, which will serve as a teacher for the distillation phase.
    \item \textbf{Phase 2: Lightweight Model Comparison.} lightweight models suitable for deployment on low-resource environments are trained and compared through two approaches: (1) Distillation: a student model is learned from the teacher model with a different loss function to adjust the knowledge transfer strategy, (2) Supervised learning: a student model is trained directly on the labeled dataset.
    \item \textbf{Phase 3: Explainability Analysis.} Performing explanation analysis using XAI techniques to investigate the model's decision-making behavior. Specifically, LIME~\citep{Ref26} is applied to generate instance-level explanations, while aggregated SHAP~\citep{Ref27} values are used to analyze feature importance at the dataset level. These methods are used to explain the model's decision-making mechanism, enhance transparency, and support the selection of a lightweight, reliable model for shrimp disease classification based on symptom descriptions.
\end{itemize}

\subsubsection{Teacher selection}
To develop a high-performance shrimp disease symptom classification model, our study fine-tuned PLMs include: 
\begin{itemize}
    \item PhoBERT~\citep{Ref03} is a Vietnamese-specific pre-trained language model developed based on the RoBERTa architecture and trained on a large-scale Vietnamese corpus with word-level tokenization. Its strong ability to capture bidirectional contextual semantics and robustness to noisy, domain-specific text make it well-suited for modeling Vietnamese shrimp disease symptom descriptions.
    \item mBART~\citep{Ref32} is a multilingual sequence-to-sequence Transformer pre-trained using a denoising auto-encoding objective across multiple languages. It is included to assess the effectiveness of a large multilingual generative model for Vietnamese shrimp disease classification and to serve as a comparative baseline against a language-specific pre-trained model.
\end{itemize}
During fine-tuning, Focal Loss~\citep{Ref33} is adopted to place greater emphasis on difficult samples and improve class discrimination during training. This loss is defined as follows:

\begin{equation}
\mathcal{L}_{focal}(p_t) = -\alpha_t (1-p_t)^{\gamma} \log(p_t)
\label{eq:focal_loss}
\end{equation}
where \(p_t\) denotes the predicted probability for the ground-truth class, \(\alpha_t\) is a class-balancing factor, and \(\gamma\) is the focusing parameter that controls the rate at which easy samples are down-weighted.
\subsubsection{Student model architecture}\label{student architecture}
\begin{figure}[H]
    \centering
    \includegraphics[width=0.4\linewidth]{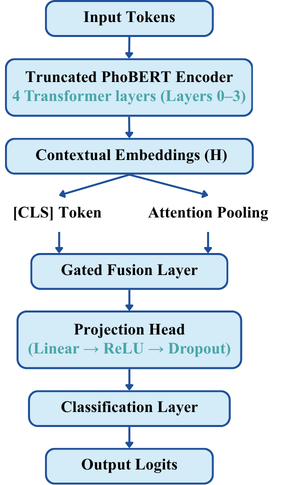}
    \caption{Architecture of the student model used in the SALT framework with a truncated PhoBERT encoder and gated fusion of contextual representations.}
    \label{fig7}
\end{figure}

The student model architecture is a lightweight variant of PhoBERT that serves as the backbone of the student network. Although the teacher model is selected from large pre-trained models such as PhoBERT or mBART, the student consistently uses a truncated PhoBERT encoder to ensure efficient inference and strong Vietnamese-language representation. To enhance contextual representation learning, we propose an architecture that incorporates a gated fusion mechanism that adaptively combines multiple contextual features. This design reduces the number of encoder layers while extending the teacher's backbone with fusion components, enabling the student model to better approximate the teacher's knowledge (Figure~\ref{fig7}). 

Specifically, the tokenized text input will be fed into a truncated PhoBERT encoder, consisting of the first four Transformer layers (Layers 0–3), to generate contextual embeddings H. From H, the model extracts two parallel representations: the \texttt{[CLS]} token embedding at the first position, representing the overall sentence semantics; and attention pooling, which aggregates token embeddings using attention weights computed by a linear layer followed by a \texttt{softmax} operation over the sequence dimension. Integrating two representations enables the model to simultaneously leverage the inherent semantic structure and adjust its weights based on context, resulting in a more profound and comprehensive representation of information. These two representations were then fused through a Gated Fusion Layer, in which a sigmoid gate controls the contribution of each information source, enabling the model to better adapt to diverse and heterogeneous contexts, particularly when pathological features are varied and complex. Next, the Fused Representation, the result of the Gated Fusion Layer, is fed into a Projection Head, which consists of a linear layer, a ReLU activation function, and dropout with a rate of $0.1$, to learn a highly separable feature space and avoid overfitting. Finally, the model employs a fully connected classification layer to predict the output logits for $10$ disease classes.

\subsubsection{Knowledge distillation}\label{KD architecture}
In this research, the student-teacher distillation method is applied to train a lighter model (the student) that inherits knowledge from a pre-trained, stronger model (the teacher). The distillation process includes two main loss components: (1) Soft Target Loss, which measures the difference in output probability distributions between the teacher and the student, where this distribution is smoothed through a softmax function with temperature \(T\), helping the student learn the relationships among closely related classes and capture semantic knowledge from the teacher; and (2) Hard Label Loss, which is calculated using the Cross-Entropy function between the student's predicted output and the ground-truth label, ensuring that the student does not deviate too far from the original target. These two components are combined using the weights \(\alpha\) for the soft target loss and \(\beta\) for the hard label loss, resulting in the general loss function shown in Equation~(\ref{eq:distillation_loss}).

\begin{equation}
\mathcal{L}_{\mathrm{distill}} = \alpha \mathcal{L}_{\mathrm{soft}} + \beta \mathcal{L}_{\mathrm{hard}}
\label{eq:distillation_loss}
\end{equation}

To comprehensively evaluate the role of loss functions in distillation, this study compares several distillation-based and representation learning losses:
\begin{itemize}
    \item Kullback--Leibler Divergence (KL)~\citep{kl}, which measures the information discrepancy between the teacher and student output probability distributions.
    \item Mean Squared Error (MSE)~\citep{mse}, which computes the squared difference between the logit vectors of the teacher and student models.
    \item Cosine Similarity loss (Cosine)~\citep{cosine}, which evaluates the angular similarity between output logit vectors, emphasizing directional agreement and capturing representational consistency between teacher and student outputs.
    \item Jensen--Shannon Divergence (JSD)~\citep{jds}, which is a symmetric and smoothed variant of KL divergence and helps improve training stability and robustness when aligning probability distributions.
    \item Supervised Contrastive (SupCon) loss~\citep{Ref34}, which optimizes the feature space by pulling samples from the same class closer while pushing samples from different classes farther apart.
    \item Center loss~\citep{Ref35}, which reduces intra-class variance by encouraging feature representations to remain close to a learned class center.
    \item Triplet loss~\citep{Ref36}, which enforces a margin between anchor--positive and anchor--negative sample pairs, thereby strengthening inter-class separability.
\end{itemize}

\subsubsection{Explainability analysis}
Beyond classification performance, the distilled student models are further evaluated using explainability analysis to better understand their decision-making mechanisms. In this study, we employ two common XAI techniques:
\begin{itemize}
    \item LIME~\citep{Ref26}, which provides local explanations by approximating the model's behavior around a given input sample using an interpretable surrogate model. It identifies the contribution of individual input features to a specific prediction, enabling fine-grained, instance-level interpretation.
    \item SHAP~\citep{Ref27}, which is grounded in cooperative game theory and computes feature importance based on Shapley values, offering a unified framework for local explanations that can be aggregated to analyze global feature importance. SHAP quantifies the marginal contribution of each feature to the model output, thereby ensuring consistency and comparability across samples.
\end{itemize}

Based on these properties, LIME and SHAP are employed to provide complementary insights at both the local and global levels, as detailed in Algorithm~\ref{alg:xai_lime_shap}. With LIME, the evaluation is performed on each sample in the test set. Specifically, LIME is applied to both the teacher and student models to extract the five most influential words together with their attribution weights. The words appearing in both explanations are identified, and the mean absolute difference between their corresponding weights is computed. If this value exceeds the threshold \(\tau\), the sample is considered to fail the explainability criterion.

\begin{algorithm}
\caption{Explainability Analysis with LIME and SHAP}
\small
\label{alg:xai_lime_shap}
\begin{algorithmic}[1]
\Require Teacher model $f_T$, Student model $f_S$, Tokenizer $T$, Test data $D_{test}$, Train data $D_{train}$
\Ensure List of unmatched samples (LIME), Global Explainability Score $E_{global}$ (SHAP)

\State Read test data $X \gets D_{test}$

\Statex \textit{// LIME -- Local Explainability}
\State Initialize LIME explainer $\mathcal{L}$ with class labels
\State Set $\texttt{unmatchedSamples} \gets \emptyset$

\Function{limePredict}{$model$, $texts$}
    \State Tokenize texts using $T$
    \State \Return softmax probabilities from $model$
\EndFunction

\For{each $x_i \in X$}
    \State $\mathcal{L}_T \gets \mathcal{L}.\texttt{explain\_instance}(x_i, \textsc{limePredict}(f_T), \texttt{num\_features}=5)$
    \State $\mathcal{L}_S \gets \mathcal{L}.\texttt{explain\_instance}(x_i, \textsc{limePredict}(f_S), \texttt{num\_features}=5)$
    \State Extract word-weight pairs $w_T$ and $w_S$
    \State $W \gets$ common words between $w_T$ and $w_S$
    \If{$W = \emptyset$}
        \State \textbf{continue}
    \EndIf
    \State $\texttt{avgDiff} \gets \mathrm{mean}(|w_T - w_S|)$
    \If{$\texttt{avgDiff} > \tau$}
        \State $\texttt{unmatchedSamples} \gets \texttt{unmatchedSamples} \cup \{(x_i,\texttt{avgDiff})\}$
    \EndIf
\EndFor

\Statex \textit{// SHAP -- Global Explainability}
\State Select background data $B \gets 200$ samples from $D_{train}$

\Function{shapPredict}{$texts$}
    \State Tokenize using $T$
    \State Pass through $f_S$ and return softmax probabilities
\EndFunction

\State Initialize SHAP explainer $\mathcal{S}$ with \textsc{shapPredict} and $B$
\State $\texttt{shapValues} \gets \mathcal{S}(X)$
\State $E_{global} \gets \mathrm{mean}(|\texttt{shapValues}|)$
\State \Return $\texttt{unmatchedSamples}$ and $E_{global}$

\end{algorithmic}
\end{algorithm}

Meanwhile, SHAP uses 200 samples from the training set as the background dataset for the explainer. SHAP values are computed on the test set, and the Global Explainability Score is calculated as the mean absolute SHAP value across all samples, where \(N\) denotes the number of test samples, as defined in Equation~(\ref{eq:global_explainability_score}). This score reflects the overall magnitude of feature contributions captured by the student model across the test set and is used to compare explainability behavior under different distillation losses.

\begin{equation}
E_{global} = \frac{1}{N} \sum_{i=1}^{N} \left| SHAP_i \right|
\label{eq:global_explainability_score}
\end{equation}

In addition to these quantitative analyses, the disease-relevant keywords identified by LIME and the globally important features highlighted by SHAP were qualitatively reviewed by a domain expert in agriculture and aquaculture. The review verified that the highlighted features were consistent with the established symptoms of the corresponding diseases, providing qualitative support for the explanation analysis.

To further assess explanation faithfulness, we perform a keyword deletion test. For each test sample, LIME is applied to the student model to extract the most positively contributing keywords associated with the predicted class. The top-$k$ keywords, where $k\in\{1,2,3,5\}$, are subsequently removed from the input sentence, and the perturbed sample is re-evaluated by the classifier. As a baseline, the same number of randomly selected words are also removed from each sample. The confidence degradation score is computed as the average reduction in the predicted probability assigned to the originally predicted class after removing the explanation-guided keywords, as defined in Equation~(\ref{eq:deletion}).

\begin{equation}
D_k=\frac{1}{N}\sum_{i=1}^{N}
\left[
p(\hat{y}_i \mid x_i)
-
p(\hat{y}_i \mid x_i \setminus S_i^{(k)})
\right]
\label{eq:deletion}
\end{equation}

where $x_i$ denotes the original input sentence, $\hat{y}_i$ is the predicted class for the original input, $S_i^{(k)}$ is the set of the top-$k$ keywords identified by LIME, and $x_i \setminus S_i^{(k)}$ represents the perturbed input obtained by removing those keywords. The function $p(\hat{y}_i \mid \cdot)$ denotes the predicted probability assigned to the originally predicted class. Larger values of $D_k$ indicate higher explanation faithfulness.

Besides confidence degradation, we also evaluate the reduction in classification accuracy after keyword removal. The accuracy degradation is computed as the difference between the original classification accuracy and the accuracy obtained after removing the top-$k$ explanation-guided keywords, as defined in Equation~(\ref{eq:acc_drop}).

\begin{equation}
\Delta Acc_k = Acc_{\mathrm{original}} - Acc_{\mathrm{deleted}}^{(k)},
\label{eq:acc_drop}
\end{equation}

where $Acc_{\mathrm{original}}$ denotes the classification accuracy on the original test set, and $Acc_{\mathrm{deleted}}^{(k)}$ denotes the classification accuracy after removing the top-$k$ keywords identified by LIME.

Finally, we compute the prediction flip rate, defined as the proportion of samples whose predicted labels change after removing the explanation-guided keywords, as given in Equation~(\ref{eq:flip}).

\begin{equation}
Flip_k=\frac{1}{N}\sum_{i=1}^{N}
\mathbb{I}\!\left(\hat{y}_i' \neq \hat{y}_i\right),
\label{eq:flip}
\end{equation}

where $\hat{y}_i$ denotes the predicted label for the original input, $\hat{y}_i'$ denotes the predicted label after removing the top-$k$ keywords, and $\mathbb{I}(\cdot)$ is the indicator function, which equals 1 if the predicted label changes and 0 otherwise. Higher values of $Flip_k$ indicate that the explanations identify words that have a stronger influence on the model's predictions.

\subsection{Experimental setup}

\subsubsection{Experimental configuration}

\textbf{Training Setup.}
To ensure fairness in performance comparison, all models are trained using identical optimization settings. The detailed hyperparameters used for fine-tuning the teacher models are summarized in Table~\ref{tab:training_hyperparameters}. Unless otherwise specified, the same configuration is applied to all models.

\begin{table}[H]
\centering
\caption{Training hyperparameters}
\label{tab:training_hyperparameters}
\begin{tabular}{lc}
\hline
\textbf{Name} & \textbf{Value} \\
\hline
Batch size & 16 \\
Epoch & 15 \\
Learning rate & 3e-5 \\
Optimizer & AdamW \\
Loss Function & Focal Loss \\
\hline
\end{tabular}
\end{table}

\textbf{Method-specific Hyperparameters.}
The hyperparameters specific to knowledge distillation, metric-learning objectives, and explainability analysis are summarized in Table~\ref{tab:method_hyperparameters}. The values are kept fixed throughout all experiments unless otherwise specified. This configuration ensures fair comparison across different distillation objectives while improving the reproducibility of the experimental results.

\begin{table}[H]
\centering
\caption{Method-specific hyperparameters used in SALT}
\label{tab:method_hyperparameters}
\begin{tabular}{lll}
\hline
\textbf{Component} & \textbf{Parameter} & \textbf{Value} \\
\hline
Distillation & $\alpha$ & 0.8 \\
Distillation & $\beta$ & 0.2 \\
Distillation & Temperature $T$ & 3.0 \\
Teacher (Focal) & $\gamma$ & 2.5 \\
Teacher (Focal) & Class weight $\alpha$ & Class-balanced \\
Center Loss & $\lambda_{\mathrm{center}}$ & 0.003 \\
SupCon & Temperature & 0.07 \\
Triplet & Margin & 0.3 \\
Triplet & Mining strategy & Batch-hard \\
LIME & Matching threshold $\tau$ & 0.15 \\
LIME & Top features $k$ & 5 \\
LIME & Perturbations & 2000 \\
\hline
\end{tabular}
\end{table}

\textbf{Evaluation Metrics.}
Accuracy (Acc) (Equation~(\ref{eq:accuracy})), Precision (Equation~(\ref{eq:precision})), Recall (Equation~(\ref{eq:recall})), and F1-Score (F1) (Equation~(\ref{eq:f1score})) are utilized as the evaluation metrics in the experiment. Meanwhile, the best model is selected based on F1 performance during the teacher phase of distillation.

\begin{equation}
\label{eq:accuracy}
Acc = \frac{TP + TN}{TP + TN + FP + FN}
\end{equation}

\begin{equation}
\label{eq:precision}
Precision = \frac{TP}{TP + FP}
\end{equation}

\begin{equation}
\label{eq:recall}
Recall = \frac{TP}{TP + FN}
\end{equation}

\begin{equation}
\label{eq:f1score}
F1 = 2 \times \frac{Precision \times Recall}{Precision + Recall}
\end{equation}

\textbf{Efficiency Measurement.}
Computational efficiency was evaluated using parameter count, model size (MB), floating-point operations per forward pass (GFLOPs), inference latency (ms/sample), and throughput (samples/s). Experiments were conducted on a workstation equipped with an NVIDIA GeForce RTX~4060 GPU (CUDA~12.4, PyTorch~2.6.0) and an Intel Xeon E5-2667 v2 CPU @ 3.30\,GHz (Windows~10). Inference was measured with batch size~1 and a maximum sequence length of 384 tokens. Each model was warmed up for 20 forward passes, and latency was averaged over 100 timed iterations. GPU latency was measured during test-set inference, whereas CPU latency and GFLOPs were obtained using an identical benchmarking protocol.

\textbf{Experimental Protocol.}
Unless otherwise specified, all experiments were repeated five times with different random seeds. The reported results are presented as the mean $\pm$ standard deviation across the five runs. Model selection was performed exclusively on the validation set using the macro-F1 score, while the test set was reserved for final evaluation.

\subsubsection{Baselines and comparative setups}
To further evaluate the effectiveness of the proposed SALT framework, we implement two categories of comparative experiments: (1) comparative training setups that analyze the contribution of individual components within the proposed framework, and (2) internal and external baseline methods for benchmarking against representative alternatives.

\textbf{Distillation with PhoBERT-L teacher.} To investigate the impact of teacher model capacity on KD performance, PhoBERT-L (370M params) is fine-tuned and employed as the teacher model in the distillation process. The student model adopts the architecture and the same distillation configuration presented in Section~\ref{student architecture}. This setup allows analysis of how a larger teacher influences the knowledge transfer process and the resulting performance of the distilled student model.

\textbf{Distilled Student without Gated Fusion.}
To evaluate the contribution of the proposed gated fusion module, an architectural ablation is conducted by removing the gated fusion mechanism from the student model. In this variant, the student retains the same lightweight backbone described in Section~\ref{student architecture}, including the truncated PhoBERT encoder, projection head, and classification layer, while replacing the gated fusion representation with the original \texttt{[CLS]} token embedding. All remaining components, including the teacher models, distillation objectives, hyperparameters, data splits, and optimization settings, are kept identical to those used in the SALT framework. This setup isolates the gated fusion module's contribution to overall classification performance.

\textbf{Training with Linguistic Augmentation.}
To evaluate the impact of linguistic data augmentation, the student model is additionally trained using the augmented training set described in Section~\ref{dataset}. The validation and test sets remain identical to those used in the proposed framework, and all remaining training configurations are kept unchanged to ensure a fair comparison. This comparative setup is included solely to assess the contribution of linguistic data augmentation.

\textbf{Supervised Fine-tuning (SFT).} To assess the effectiveness of KD, an internal baseline is established by directly training the student model on the labeled dataset without distillation. The student's architecture follows the design described in Section~\ref{student architecture}. To ensure a fair comparison, the model is trained with the same data split and optimization settings as the teacher models. Early stopping based on validation macro-F1 is used to prevent overfitting due to the student model's reduced capacity.

\textbf{TF-IDF features and a deep neural network.} An external baseline based on TF-IDF features and a deep neural network (DNN) is reimplemented following the configuration reported in ~\citep{Ref19}. Among the configurations reported by the original study, only the best-performing TF-IDF + DNN model is reproduced for comparison. Specifically, each symptom description is represented using word-level TF-IDF features with bi-gram representation, with threshold parameters set to $min_{df}=2$ and $max_{df}=0.95$. The obtained vectors are then fed into a two-layer feedforward neural network (Dense $1024$ $\rightarrow$ Dense $512$, with ReLU activation function and dropout), and are finalized with a softmax layer for disease class prediction.

\textbf{F2LLM-v2-80M.}
As an additional external baseline, we employ F2LLM-v2-80M (\(\sim\)80M parameters), following the embedding model described in~\citep{Ref37}. It is a lightweight multilingual text embedding model from the CodeFuse F2LLM-v2 family. Each shrimp disease description is encoded into a fixed-dimensional sentence representation using the pre-trained encoder, where the $\ell_2$-normalized hidden state of the end-of-sequence token is used as the sentence embedding. The encoder is kept frozen throughout training, while a two-layer feed-forward classifier (512 $\rightarrow$ 256, dropout = 0.3) is trained on top of the extracted embeddings for ten-class single-label classification. To ensure a fair comparison, the same train, validation, and test splits as the proposed framework are adopted. The classification head is optimized using Adam~\citep{adam} with early stopping based on the validation macro-F1 score, and the checkpoint with the highest validation macro-F1 is selected for evaluation on the test set.

\section{Experimental results}
\subsection{Evaluation of teacher models}

Table~\ref{tab:teacher_results} presents the classification performance of the fine-tuned teacher models on the validation set. PhoBERT outperformed mBART, achieving an accuracy of 89.08\% and an F1-score of 0.8899, compared with 86.43\% accuracy and 0.8647 F1-score for mBART. In addition, PhoBERT required an average inference time of 13.34 ms per sample, whereas mBART required 342.16 ms, corresponding to approximately a 26$\times$ reduction in inference latency. Overall, PhoBERT achieved higher classification performance while requiring substantially fewer computational resources than mBART with its smaller parameter count (135.01 M vs. 611.91 M) and model size (515.03 MB vs. 2334.32 MB).

\begin{table}
\centering
\small
\caption{Evaluation results of fine-tuned teacher models on the shrimp disease classification validation set.}
\label{tab:teacher_results}
\begin{tabular}{lrr}
\hline
\textbf{Metrics} & \textbf{PhoBERT-base} & \textbf{mBART} \\
\hline
Acc (\%) & 89.08 $\pm$ 1.71 & 86.43 $\pm$ 2.03 \\
F1 & 0.8899 $\pm$ 0.0182 & 0.8647 $\pm$ 0.0200 \\
Precision & 0.8948 $\pm$ 0.0163 & 0.8691 $\pm$ 0.0193 \\
Recall & 0.8903 $\pm$ 0.0175 & 0.8637 $\pm$ 0.0203 \\ 
Inference Time (ms) & 13.34 $\pm$ 0.64 & 342.16 $\pm$ 61.21 \\\hline
Params (M) & 135.01 & 611.91 \\
Size (MB) & 515.03 & 2334.32 \\
\hline
\end{tabular}
\end{table}

\begin{figure}[H]
    \centering
    \includegraphics[width=1\linewidth]{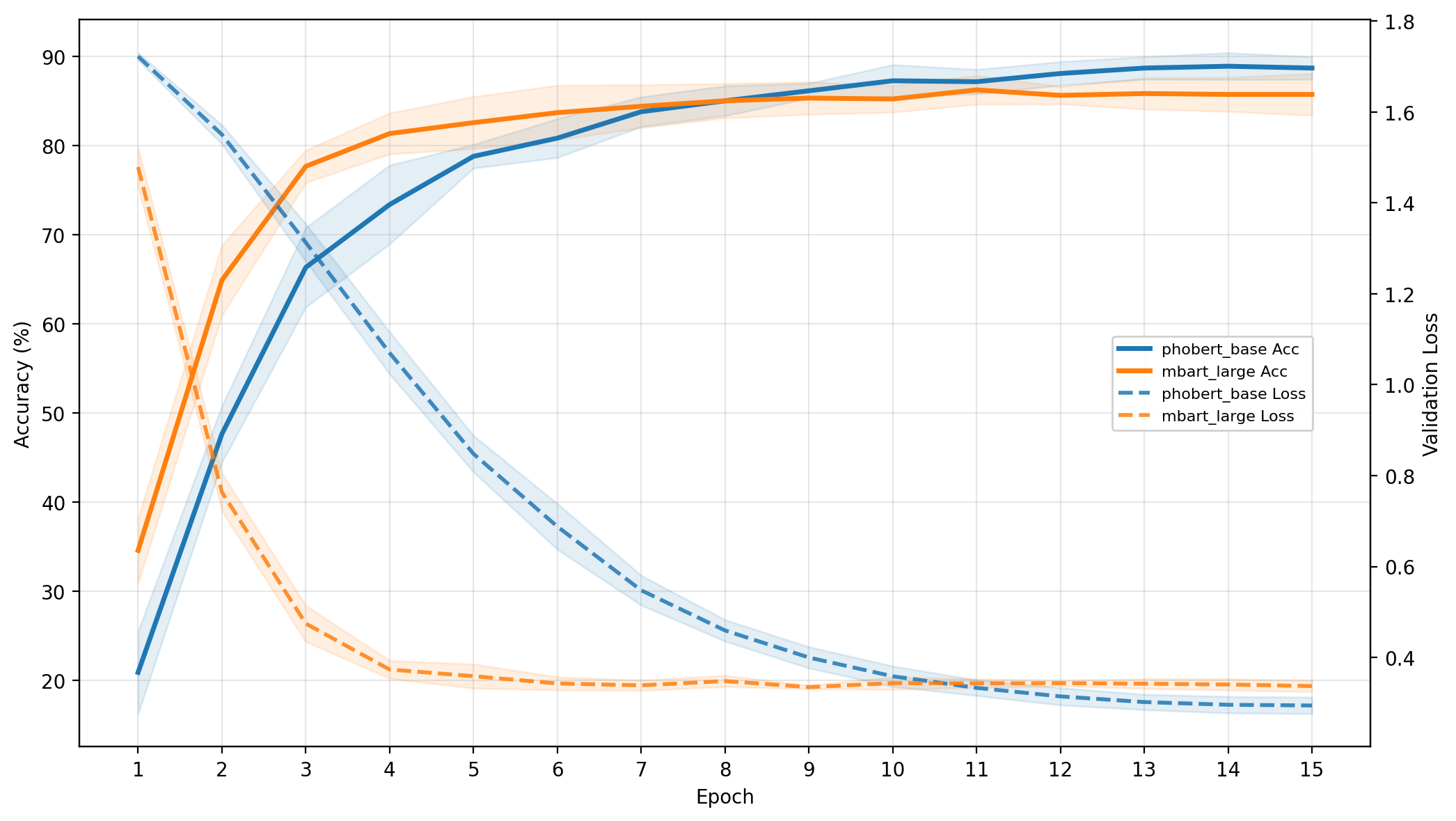}
    \caption{Accuracy and loss curves on the validation set for the two teacher models: PhoBERT-base and mBART.}
    \label{fig8}
\end{figure}

As illustrated in Figure~\ref{fig8}, the training dynamics of PhoBERT and mBART are presented as the mean and standard deviation over five runs with different random seeds. PhoBERT exhibits a steady improvement in accuracy, increasing from approximately 21\% in the first epoch to nearly 90\% by epoch 10, after which the performance stabilizes with only minor fluctuations. Correspondingly, the loss decreases consistently throughout training, indicating stable convergence. The relatively small gap between the best and worst runs in the later epochs further demonstrates the robustness of the training process. In contrast, mBART achieves a rapid increase in accuracy during the first few epochs, rising from approximately 35\% to over 80\% by epoch 3, and then gradually stabilizes at around 86–88\% after epoch 8. Its loss decreases sharply during the initial training stage and approaches zero after approximately six epochs, suggesting fast optimization. However, despite its faster early convergence, mBART ultimately achieves a lower final accuracy than PhoBERT.

\begin{figure}[H]
    \centering
    \includegraphics[width=1\linewidth]{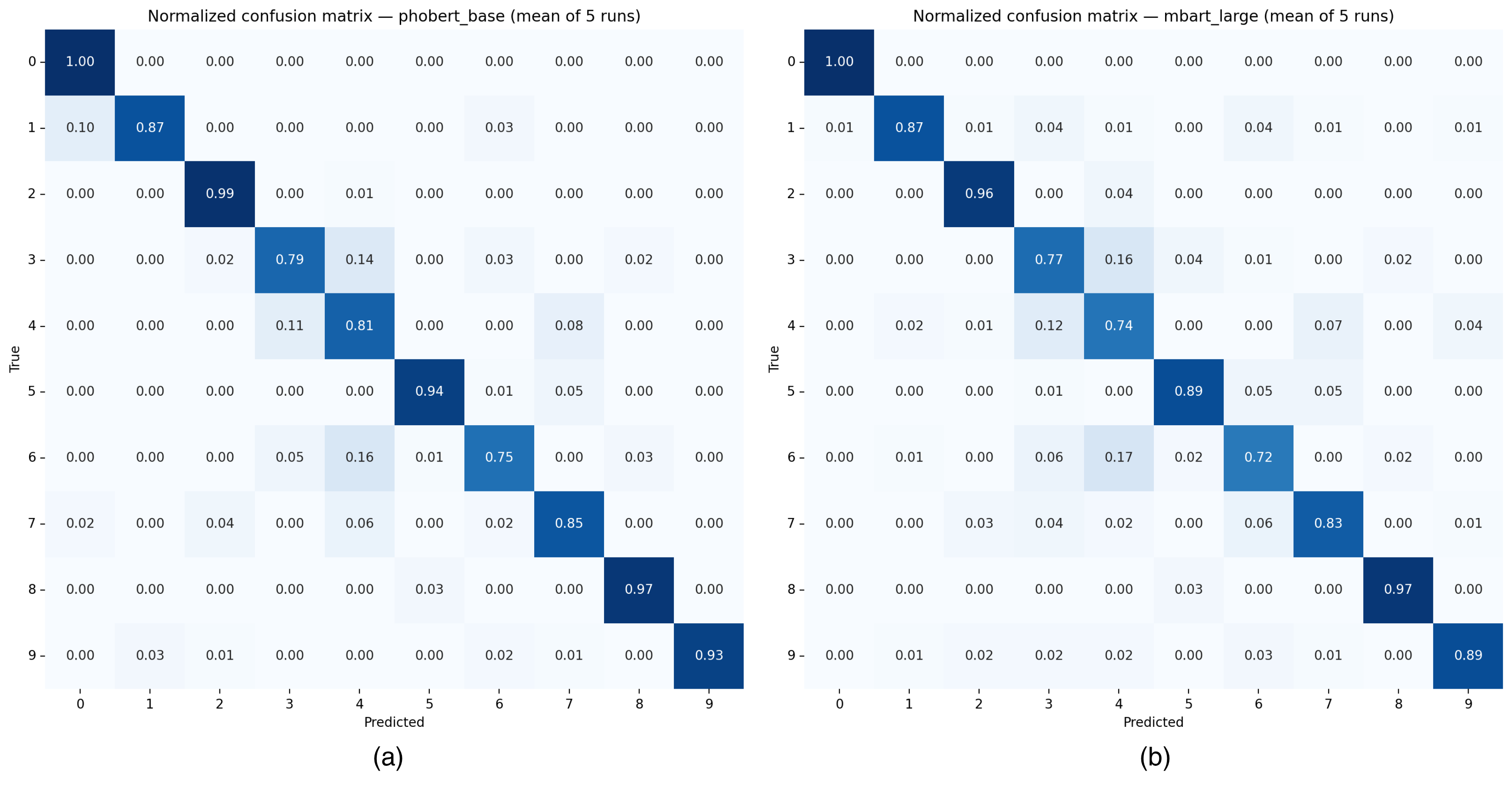}
    \caption{Normalized confusion matrices after fine-tuning on the shrimp disease classification dataset for two models: PhoBERT and mBART. Class indices: 0 = Acute Hepatopancreatic Necrosis Disease; 1 = White Feces Disease; 2 = White Spot Disease; 3 = Luminous Bacteria Disease; 4 = Loose Shell Syndrome; 5 = Black Gill Disease; 6 = Filamentous Bacterial Disease; 7 = Vitamin C Deficiency Disease; 8 = Yellow Head Disease; 9 = Taura Syndrome.}
    \label{fig9}
\end{figure}

Figure~\ref{fig9} presents the normalized confusion matrices of the fine-tuned PhoBERT and mBART models. Both models achieve high classification performance across most disease classes, as indicated by the dominant diagonal values. Overall, PhoBERT demonstrates better class discrimination than mBART, achieving higher correct classification rates for several challenging disease categories, particularly Loose Shell Syndrome (81\% vs. 74\%), Black Gill Disease (94\% vs. 89\%), and Taura Syndrome (93\% vs. 89\%). Both models exhibit similar confusion patterns for diseases with overlapping clinical symptoms. The most prominent misclassifications occur between Luminous Bacteria Disease and Loose Shell Syndrome, as well as between Filamentous Bacterial Disease and Loose Shell Syndrome, reflecting the similarity of their symptom descriptions. Overall, PhoBERT produces fewer off-diagonal misclassifications and exhibits better class separability than mBART. Together with the quantitative performance and efficiency results, these findings support the selection of PhoBERT as the teacher model for the subsequent knowledge distillation framework.

\subsection{Evaluation of student models}

The classification performance of the student models distilled from the PhoBERT teacher using different distillation objectives is presented in Table~\ref{tab:student_loss_results}. The conventional distillation losses (KL, MSE, Cosine, and JSD) achieved comparable performance, with accuracy ranging from 87.65\% to 89.18\% and F1-scores from 0.8773 to 0.8935. Among them, Cosine loss achieved the best overall performance, obtaining the highest accuracy (89.18\%), F1-score (0.8935), precision (0.9051), and recall (0.8925). These results suggest that preserving the angular similarity between teacher and student representations is an effective strategy for transferring semantic knowledge in the shrimp disease classification task. In contrast, the metric learning-based losses (SupCon, Center, and Triplet) generally yielded lower classification performance. SupCon produced the lowest results, with an accuracy of 86.02\% and an F1-score of 0.8601, while Center and Triplet achieved comparable performance, with F1-scores of 0.8743 and 0.8737, respectively. These findings indicate that metric learning objectives do not provide additional benefits over conventional distillation losses for this task. Regarding computational efficiency, all distilled student models exhibited similar inference times, ranging from 3.75 to 3.83 ms per sample, with Triplet loss achieving the lowest latency (3.75 ms). Since all student models shared the same architecture, the differences in inference time across distillation objectives were negligible.

\begin{table}
\centering
\scriptsize
\setlength{\tabcolsep}{3pt}
\caption{Performance of student models distilled from PhoBERT-base using different distillation objectives on the test set (The best indicators are in bold).}
\label{tab:student_loss_results}
\begin{tabular}{lccccc}
\hline
\textbf{Metric} & \textbf{Acc (\%)} & \textbf{F1} & \textbf{Precision} & \textbf{Recall} & \textbf{Time (ms)} \\ 
\hline
\textbf{KL}                                    & 88.67($\pm$2.17)                       & 0.8880 ($\pm$0.0213)             & 0.8997 ($\pm$0.0172)                    & 0.8872($\pm$0.0219)                  & 3.78($\pm$0.11)                                   \\ 
\textbf{MSE}                                   & 88.16 ($\pm$1.22)                          & 0.8822 ($\pm$0.0134)                 & 0.8940 ($\pm$0.0144)                        & 0.8818 ($\pm$0.0130)                     & 3.77 ($\pm$0.04)                                      \\ 
\textbf{JSD}                                   & 87.65 ($\pm$2.21)                          & 0.8773 ($\pm$0.021)                  & 0.8882 ($\pm$0.0191)                        & 0.8768 ($\pm$0.0219)                     & 3.78 ($\pm$0.09)                                      \\ 
\textbf{SupCon}                                & 86.02 ($\pm$1.23)                          & 0.8601 ($\pm$0.0138)                 & 0.8758 ($\pm$0.0103)                        & 0.8601 ($\pm$0.0135)                     & 3.77 ($\pm$0.06)                                      \\ 
\textbf{Center}                                & 87.45 ($\pm$2.31)                          & 0.8743 ($\pm$0.0231)                 & 0.8810 ($\pm$0.0247)                        & 0.8745 ($\pm$0.023)                       & 3.80 ($\pm$0.09)                                      \\ 
\textbf{Triplet}                               & 87.35 ($\pm$1.22)                          & 0.8737 ($\pm$0.0138)                 & 0.8830 ($\pm$0.0141)                        & 0.8738 ($\pm$0.0125)                     & \textbf{3.75 ($\pm$0.02)}                             \\ 
\textbf{Cosine}                                & \textbf{89.18 ($\pm$1.76)}                 & \textbf{0.8935 ($\pm$0.0173)}         & \textbf{0.9051 ($\pm$0.0163)}               & \textbf{0.8925 ($\pm$0.0179)}            & 3.83 ($\pm$0.15)                                      \\ \hline
\end{tabular}
\end{table}

\begin{figure}
    \centering
    \includegraphics[width=1\linewidth]{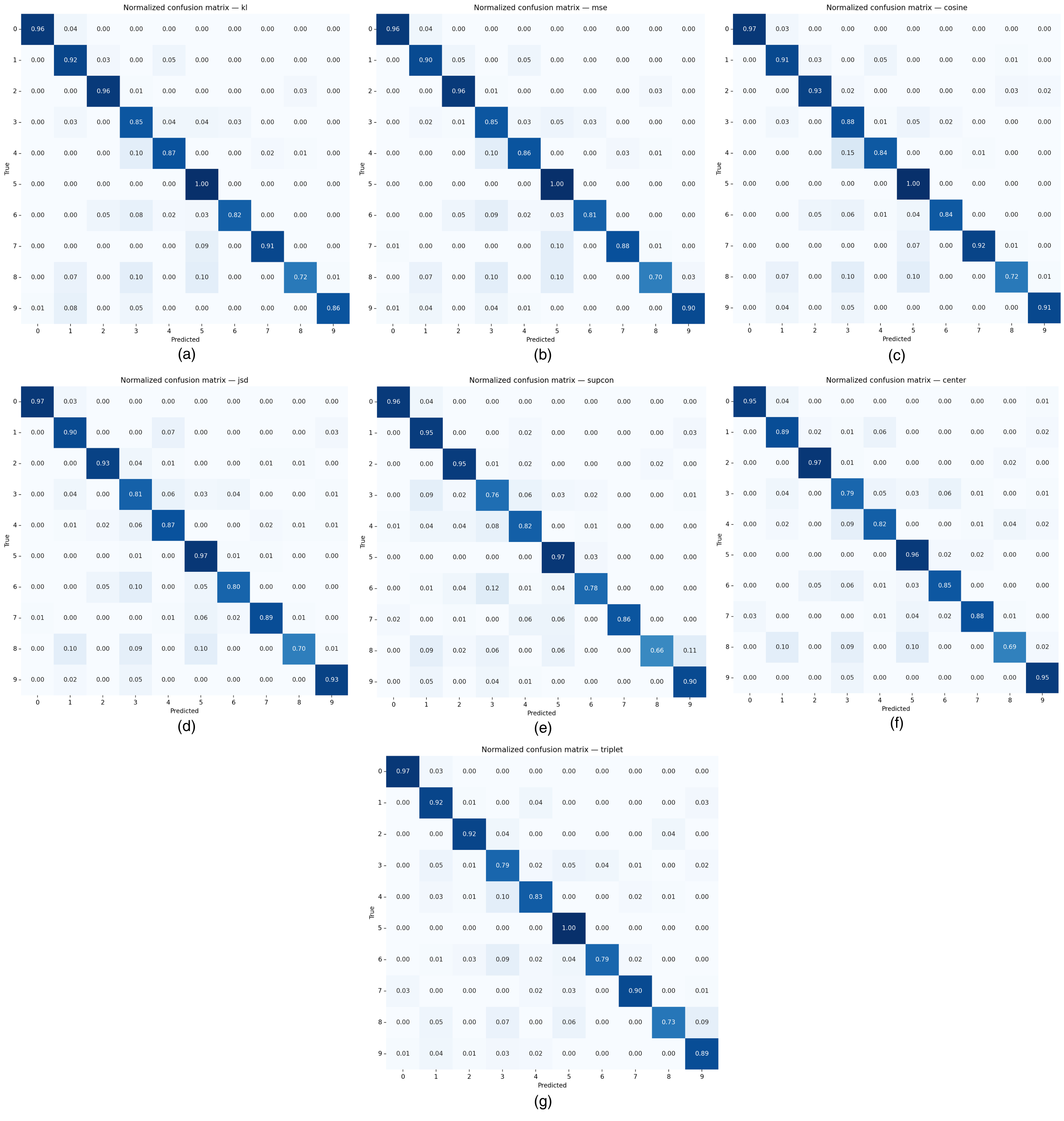}
    \caption{Normalized confusion matrices for student models distilled from PhoBERT-base using different loss functions: KL, MSE, Cosine, JSD, SupCon, Center, and Triplet. Class indices: 0 = Acute Hepatopancreatic Necrosis Disease; 1 = White Feces Disease; 2 = White Spot Disease; 3 = Luminous Bacteria Disease; 4 = Loose Shell Syndrome; 5 = Black Gill Disease; 6 = Filamentous Bacterial Disease; 7 = Vitamin C Deficiency Disease; 8 = Yellow Head Disease; 9 = Taura Syndrome.}
    \label{fig10}
\end{figure}

Figure~\ref{fig10} presents the normalized confusion matrices of the student models distilled using different distillation objectives. Overall, the conventional distribution-based objectives (KL, MSE, Cosine, and JSD) exhibit stronger diagonal dominance and more consistent class discrimination than the metric learning-based objectives. Among them, Cosine loss produces the cleanest confusion matrices, achieving the highest correct classification rates for several challenging disease categories, including Yellow Head Disease (72\%) and Loose Shell Syndrome (84\%). Across all objectives, the most frequent misclassifications occur among Luminous Bacteria Disease, Loose Shell Syndrome, Filamentous Bacterial Disease, and Yellow Head Disease, reflecting the overlap in their symptom descriptions. In contrast, diseases such as Acute Hepatopancreatic Necrosis Disease and Black Gill Disease remain well separated, with diagonal values consistently of at least 95\%. Compared with the conventional objectives, the metric learning-based objectives produce more dispersed off-diagonal errors, particularly SupCon, whereas Center and Triplet provide intermediate performance. Overall, the confusion matrices analysis corroborates the quantitative results, further demonstrating the effectiveness of conventional distribution-based distillation objectives, especially Cosine loss.

\subsection{Comparison with baseline methods.} \textbf{Supervised Fine-tuning (internal baseline)}
The directly supervised student model achieves an accuracy of 86.53\% and an F1-score of 0.8666, as reported in Table~\ref{tab:tabVII}. Compared with the distilled student models (Table~\ref{tab:student_loss_results}), supervised fine-tuning consistently yields lower classification performance. These results indicate that knowledge distillation effectively transfers semantic knowledge from the teacher model, leading to improved predictive performance over conventional supervised training while maintaining the same student architecture.

\begin{table}
\centering
\scriptsize
\setlength{\tabcolsep}{4pt}
\caption{Performance comparison of baseline models on the test set (The best indicators are in bold).}
\label{tab:tabVII}
\begin{tabular}{lccc}
\hline
\textbf{Metric} & \textbf{SFT} & \textbf{TF-IDF + DNN} & \textbf{F2LLM-v2-80M} \\
\hline

Acc (\%) & 86.53 ($\pm$2.44) & \textbf{88.98} ($\pm$0.72) & 82.86 ($\pm$2.03) \\

F1 & 0.8666 ($\pm$0.0250) & \textbf{0.8896} ($\pm$0.0071) & 0.8287 ($\pm$0.0201) \\

Precision & 0.8803 ($\pm$0.0280) & \textbf{0.9021} ($\pm$0.0062) & 0.8367 ($\pm$0.0193) \\

Recall & 0.8659 ($\pm$0.0246) & \textbf{0.8896} ($\pm$0.0071) & 0.8297 ($\pm$0.0200) \\

Inference Time (ms) & \textbf{3.91} ($\pm$0.04) & 76.46 ($\pm$0.44) & 5.53 ($\pm$0.13) \\
\hline
\end{tabular}
\end{table}

\begin{figure}[H]
    \centering
    \includegraphics[width=0.8\linewidth]{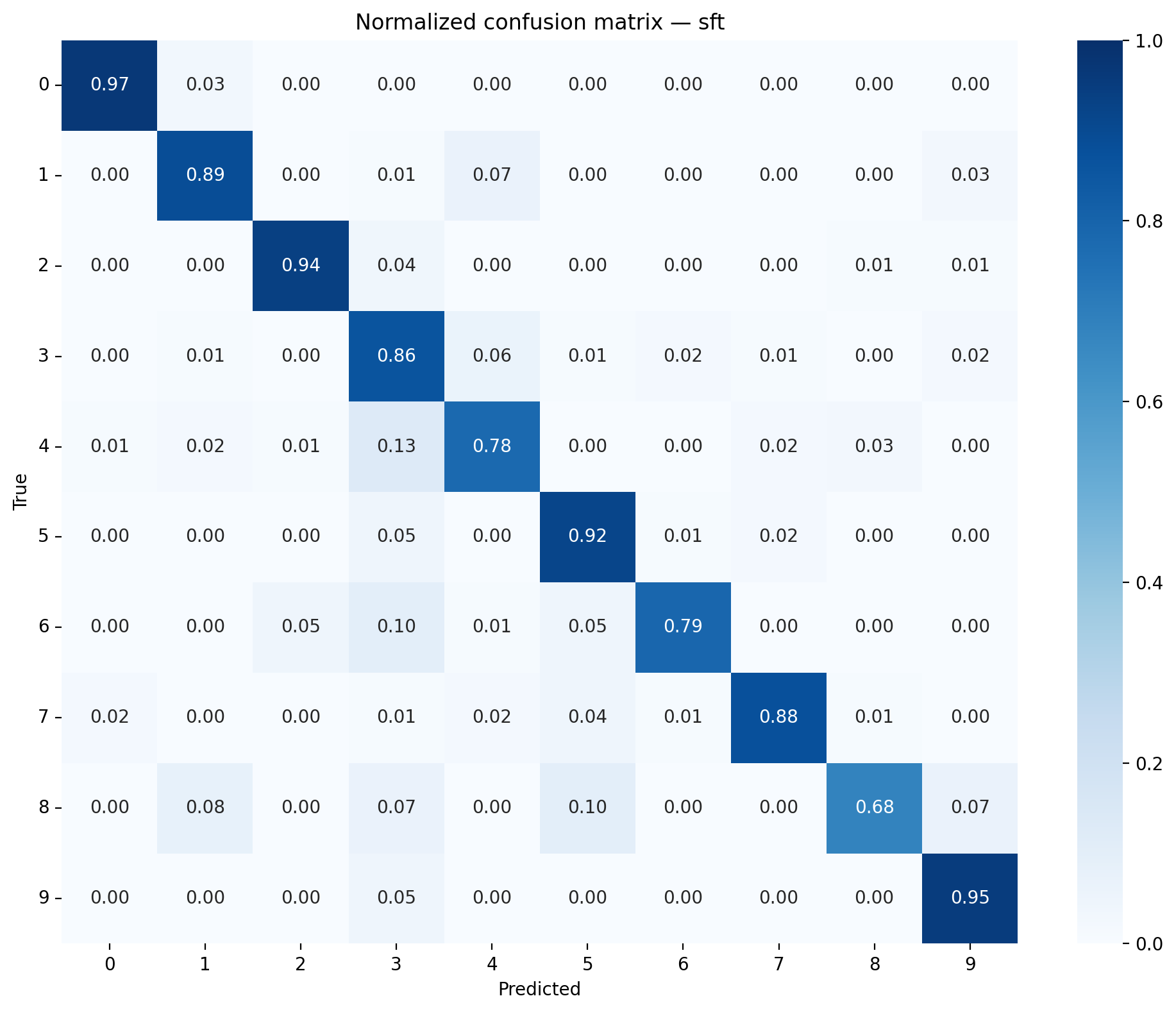}
    \caption{Normalized confusion matrices of the student model trained with supervised fine-tuning. Class indices: 0 = Acute Hepatopancreatic Necrosis Disease; 1 = White Feces Disease; 2 = White Spot Disease; 3 = Luminous Bacteria Disease; 4 = Loose Shell Syndrome; 5 = Black Gill Disease; 6 = Filamentous Bacterial Disease; 7 = Vitamin C Deficiency Disease; 8 = Yellow Head Disease; 9 = Taura Syndrome.}
    \label{fig15}
\end{figure}

The confusion patterns of the supervised fine-tuning baseline, shown in Figure~\ref{fig15}, reveal that the model distinguishes well between disease classes with distinctive symptom descriptions while exhibiting increased confusion among clinically similar diseases. The model achieves high correct classification rates for several well-separated disease classes, including Acute Hepatopancreatic Necrosis Disease (97\%), Taura Syndrome (95\%), and White Spot Disease (94\%). In contrast, lower classification rates are observed for Loose Shell Syndrome (78\%), Filamentous Bacterial Disease (79\%), and particularly Yellow Head Disease (68\%), which exhibit more frequent confusion with disease classes sharing similar clinical symptom descriptions. The supervised fine-tuning baseline exhibits more off-diagonal errors for these challenging disease categories, suggesting that knowledge distillation helps reduce inter-class confusion and enables the student model to learn more discriminative semantic representations.

\textbf{TF-IDF features and a deep neural network (external baseline).}As shown in Table~\ref{tab:tabVII}, the TF-IDF + DNN baseline achieves an accuracy of 88.98\% and an F1-score of 0.8896, outperforming both the supervised fine-tuning baseline and the F2LLM-v2-80M baseline. Its classification performance is competitive with several knowledge distillation variants but remains slightly lower than the best-performing distilled student model using Cosine loss (89.18\% accuracy and 0.8935 F1-score). Although the TF-IDF + DNN baseline provides strong predictive performance, it requires substantially longer inference time (76.46 ms per sample) than the neural student models, limiting its computational efficiency.

\begin{figure}[H]
    \centering
    \includegraphics[width=0.6\linewidth]{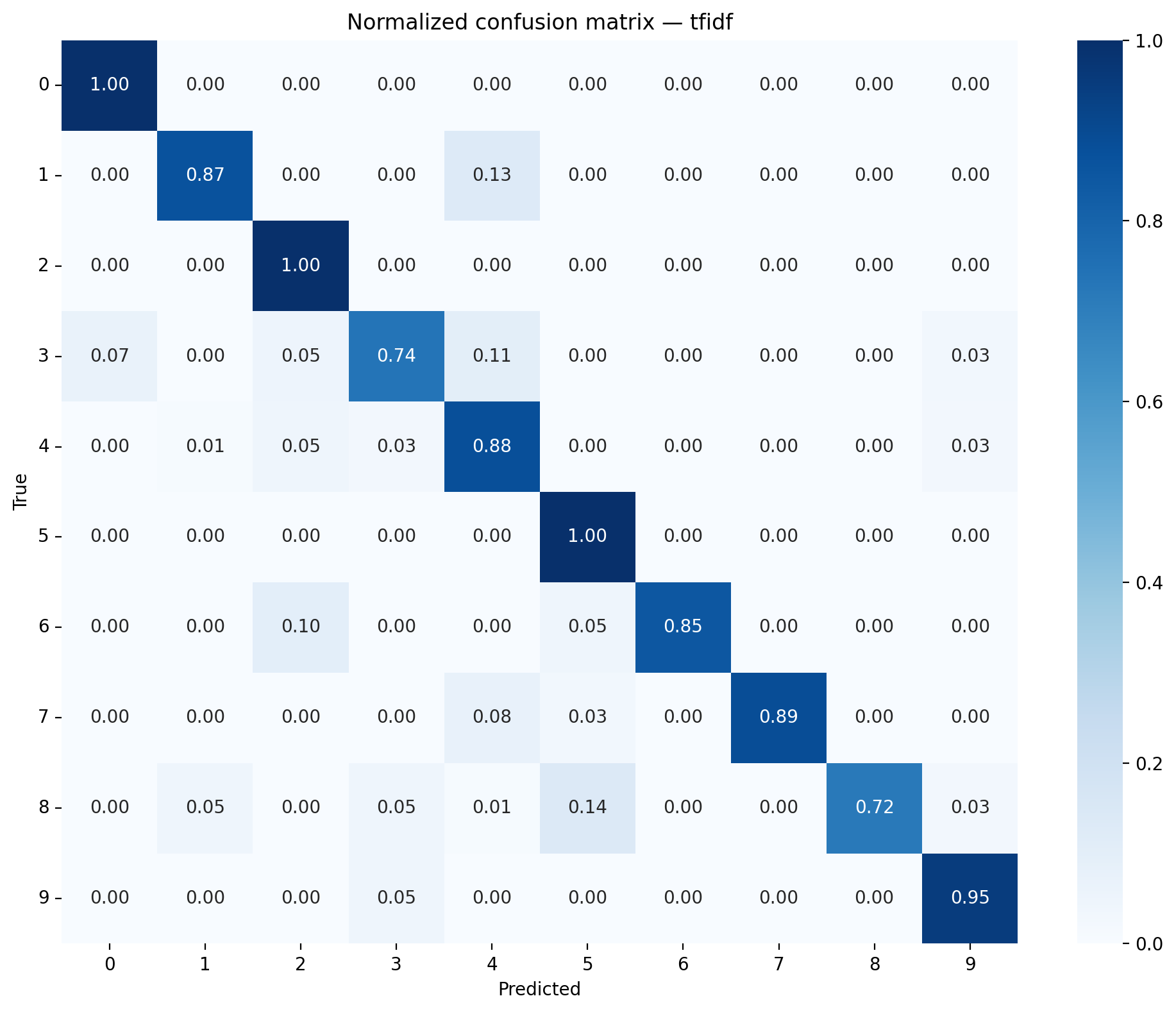}
    \caption{Normalized confusion matrices of the TF-IDF + DNN reference model. Class Indices: 0 = Acute Hepatopancreatic Necrosis Disease; 1 = White Feces Disease; 2 = White Spot Disease; 3 = Luminous Bacteria Disease; 4 = Loose Shell Syndrome; 5 = Black Gill Disease; 6 = Filamentous Bacterial Disease; 7 = Vitamin C Deficiency Disease; 8 = Yellow Head Disease; 9 = Taura Syndrome.}
    \label{fig16}
\end{figure}

As illustrated in Figure~\ref{fig16}, the TF-IDF + DNN baseline exhibits strong class discrimination for several disease categories, achieving perfect classification for Acute Hepatopancreatic Necrosis Disease, White Spot Disease, and Black Gill Disease. In contrast, lower classification rates are observed for Luminous Bacteria Disease (74\%) and Yellow Head Disease (72\%), which are more frequently confused with diseases sharing similar clinical symptom descriptions. These confusion patterns suggest that although TF-IDF features effectively capture discriminative lexical cues, their sparse bag-of-words representation is less effective at modeling semantic similarity among clinically related diseases.

\textbf{F2LLM-v2-80M}
Table~\ref{tab:tabVII} summarizes the performance of the F2LLM-v2-80M baseline. The model achieves an accuracy of 82.86\% and an F1-score of 0.8287, representing the lowest classification performance among the evaluated baseline methods. Compared with the TF-IDF + DNN baseline and the supervised fine-tuning baseline, F2LLM-v2-80M yields lower accuracy, precision, recall, and F1-score. In addition, its inference time (5.53 ms per sample) is higher than that of the supervised fine-tuning baseline (3.91 ms per sample). 

\begin{figure}[H]
    \centering
    \includegraphics[width=0.8\linewidth]{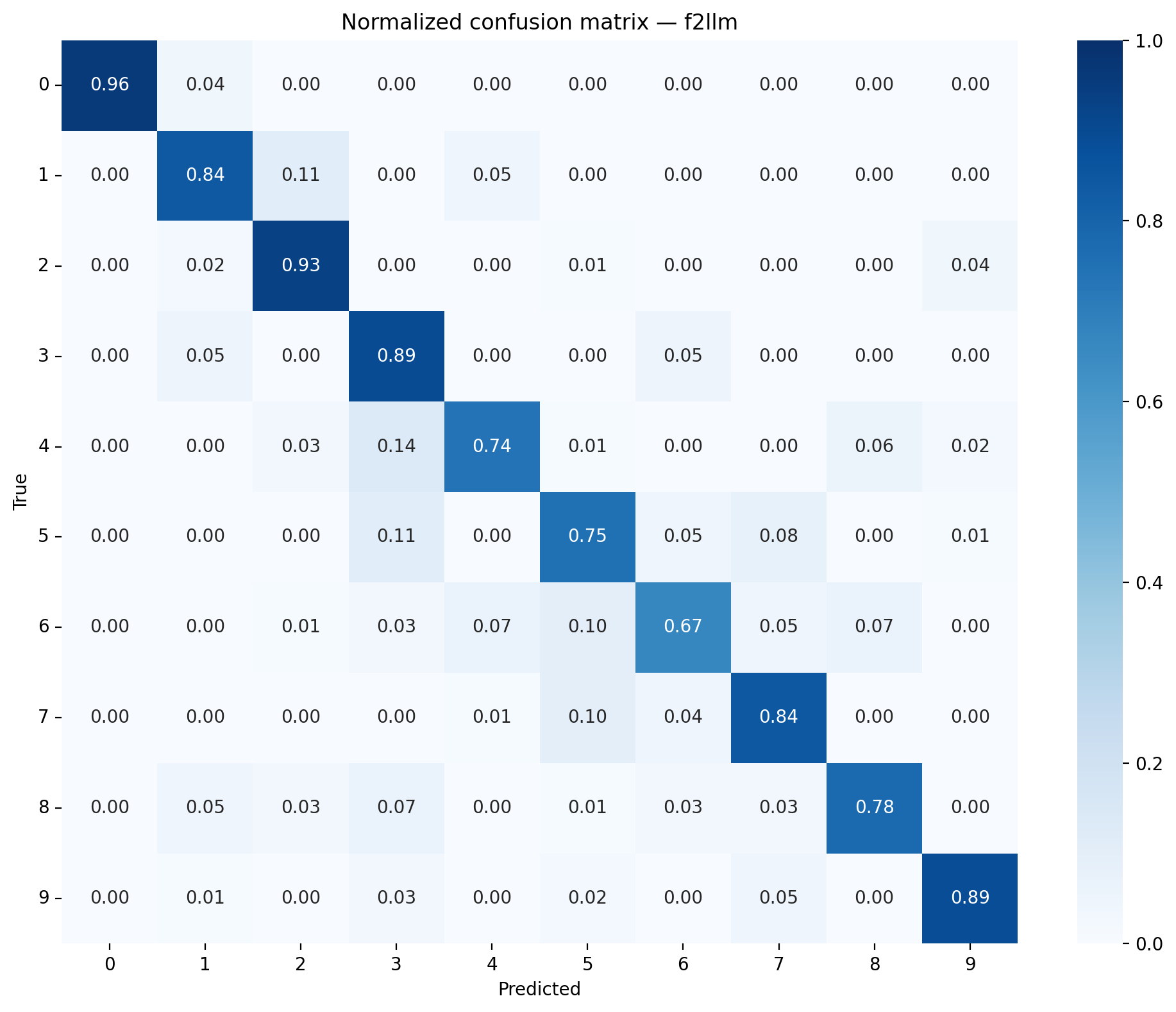}
    \caption{Normalized confusion matrix of the F2LLM-v2-80M baseline. Class indices: 0 = Acute Hepatopancreatic Necrosis Disease; 1 = White Feces Disease; 2 = White Spot Disease; 3 = Luminous Bacteria Disease; 4 = Loose Shell Syndrome; 5 = Black Gill Disease; 6 = Filamentous Bacterial Disease; 7 = Vitamin C Deficiency Disease; 8 = Yellow Head Disease; 9 = Taura Syndrome.}
    \label{fig17}
\end{figure}

The confusion matrix of the F2LLM-v2-80M baseline, shown in Figure~\ref{fig17}, indicate that the model achieves reliable classification across several well-separated disease classes, including Acute Hepatopancreatic Necrosis Disease (96\%) and White Spot Disease (93\%). However, notably lower correct classification rates are observed for Filamentous Bacterial Disease (67\%), Loose Shell Syndrome (74\%), and Black Gill Disease (75\%), which are more frequently confused with diseases sharing similar clinical symptom descriptions. In particular, Loose Shell Syndrome is frequently misclassified as Luminous Bacteria Disease (14\%). Overall, the confusion matrix reveals weaker class separability than the proposed knowledge distillation framework.

\subsection{Ablation study}
\subsubsection{Effectiveness of teacher model scale}
\begin{figure}
    \centering
    \includegraphics[width=0.8\linewidth]{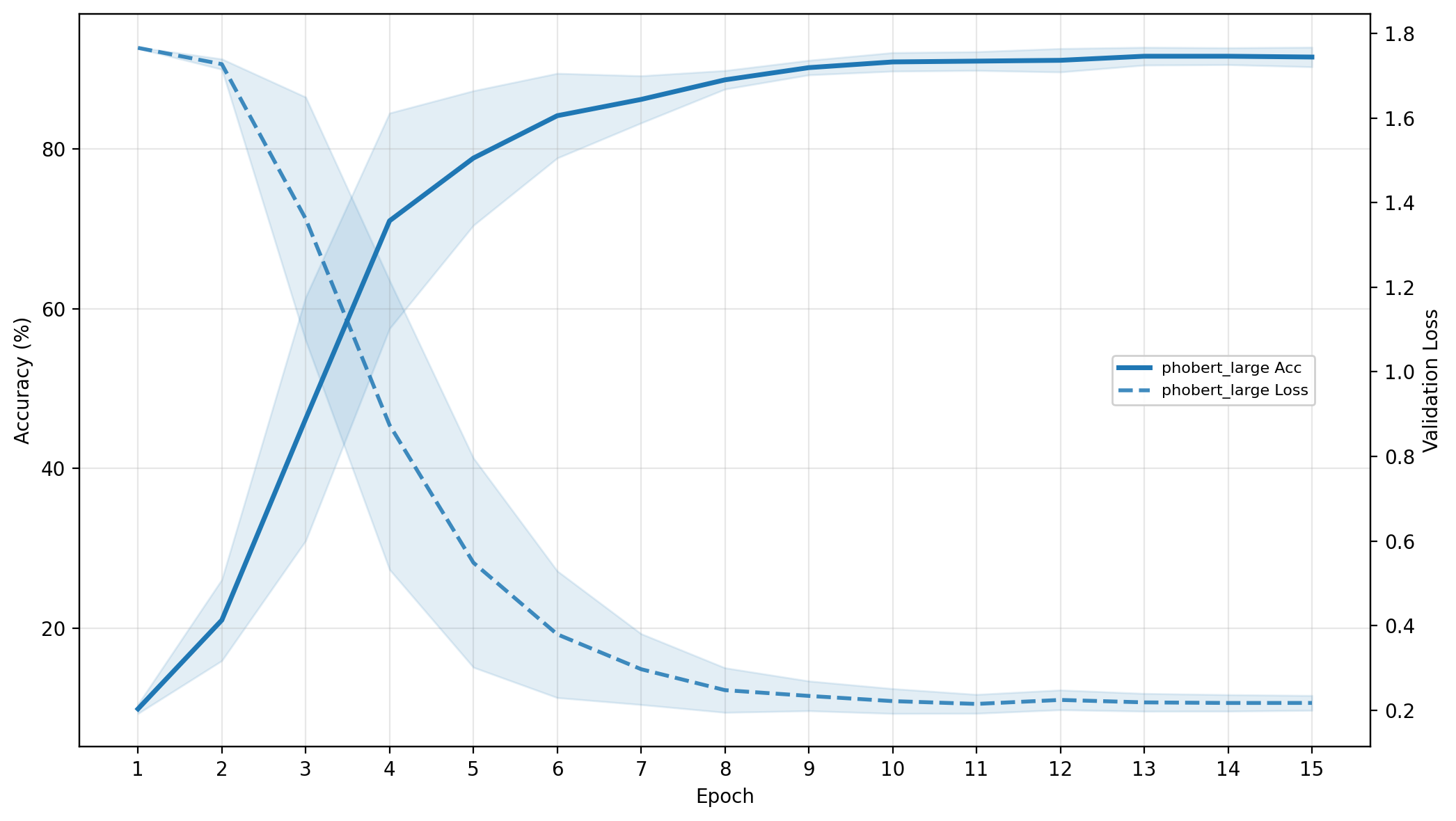}
    \caption{Accuracy and Loss Curves on the Validation Set for PhoBERT-large.}
    \label{fig11}
\end{figure}

\begin{figure}
    \centering
    \includegraphics[width=0.8\linewidth]{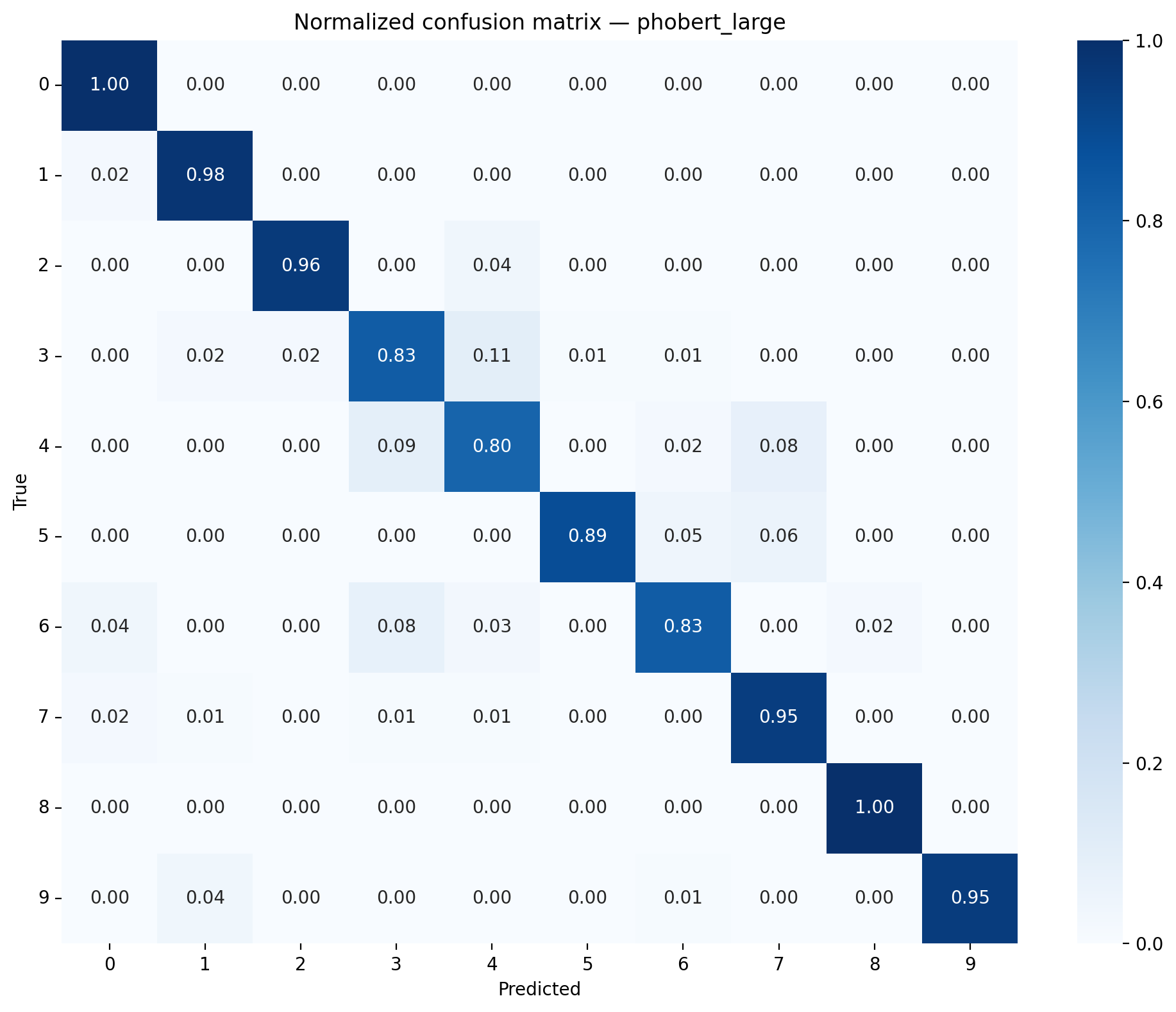}
    \caption{Normalized confusion matrices after fine-tuning on the shrimp disease classification dataset for PhoBERT-large. Class indices: 0 = Acute Hepatopancreatic Necrosis Disease; 1 = White Feces Disease; 2 = White Spot Disease; 3 = Luminous Bacteria Disease; 4 = Loose Shell Syndrome; 5 = Black Gill Disease; 6 = Filamentous Bacterial Disease; 7 = Vitamin C Deficiency Disease; 8 = Yellow Head Disease; 9 = Taura Syndrome.}   
    \label{fig12}
\end{figure}

The fine-tuned PhoBERT-large achieves an accuracy of $91.94 \pm 0.94\%$ and an F1-score of $0.9183 \pm 0.0097$, achieving slightly higher accuracy and F1-score than the PhoBERT-base teacher. The validation accuracy and loss curves (Figure~\ref{fig11}) and the confusion matrix (Figure~\ref{fig12}) indicate stable convergence and strong class-level discrimination across shrimp disease categories.

\begin{figure}
    \centering
    \includegraphics[width=1\linewidth]{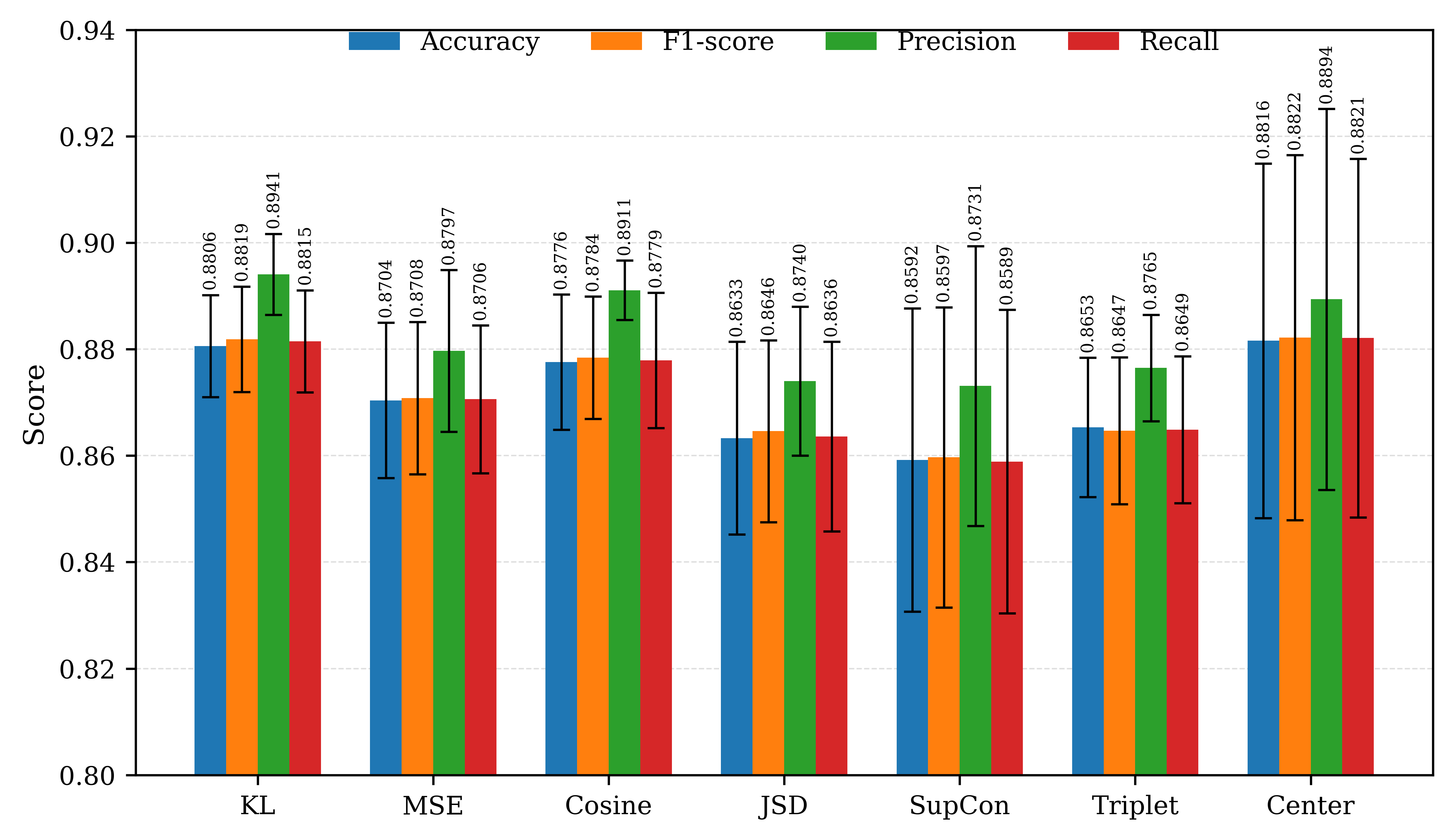}
    \caption{Performance of student models distilled from PhoBERT-large using different distillation objectives on the test set.}
    \label{figx13}
\end{figure}

\begin{figure}
    \centering
    \includegraphics[width=1\linewidth]{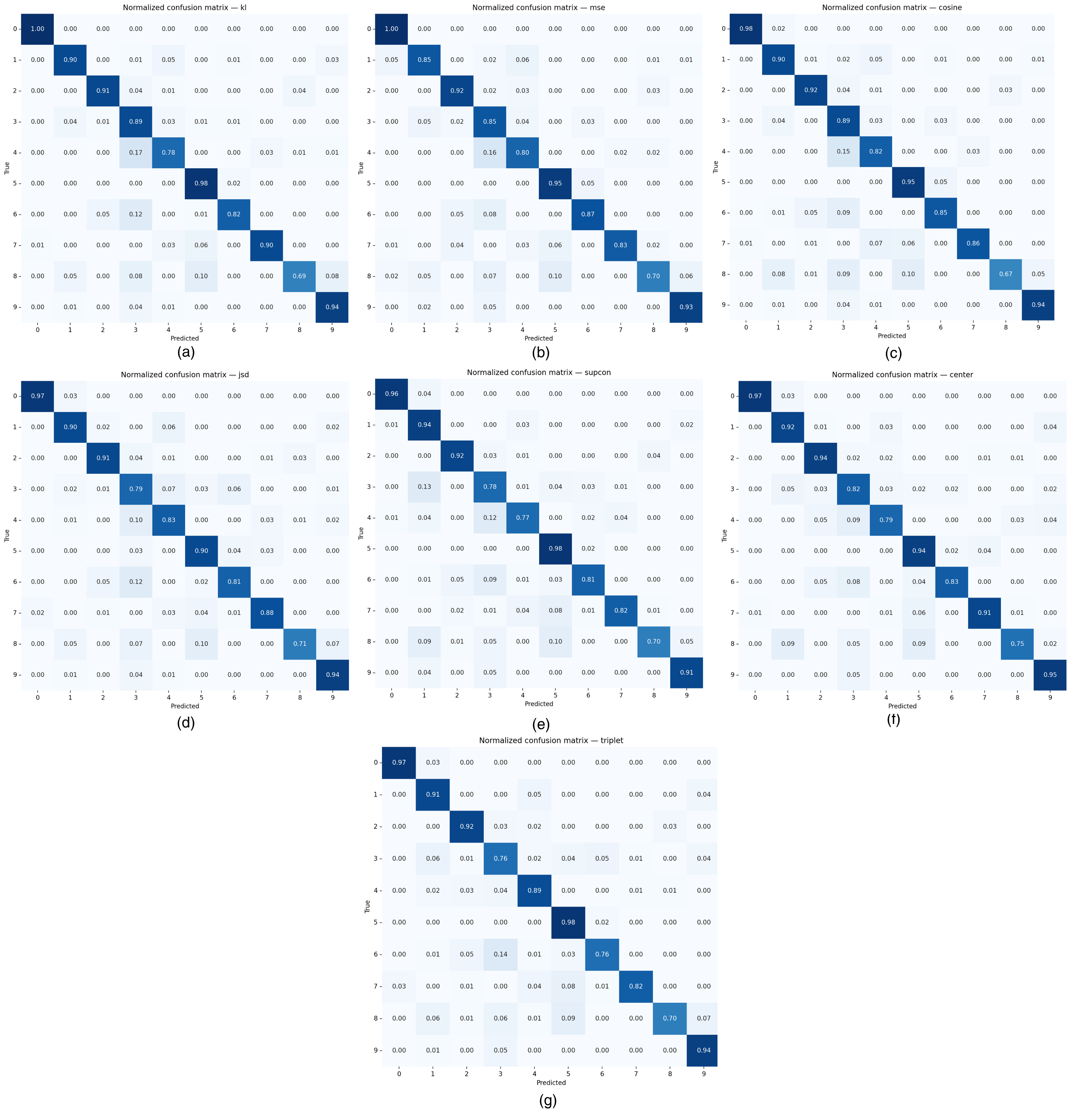}
    \caption{Normalized confusion matrices for student models distilled from PhoBERT-large using different distillation objectives: KL, MSE, Cosine, JSD, SupCon, Center, and Triplet. Class Indices: 0 = Acute Hepatopancreatic Necrosis Disease; 1 = White Feces Disease; 2 = White Spot Disease; 3 = Luminous Bacteria Disease; 4 = Loose Shell Syndrome; 5 = Black Gill Disease; 6 = Filamentous Bacterial Disease; 7 = Vitamin C Deficiency Disease; 8 = Yellow Head Disease; 9 = Taura Syndrome.}
    \label{fig13}
\end{figure}

Table~\ref{tab:dis_result_phobert_large} presents the performance of student models distilled from PhoBERT-large using different distillation objectives on the test set. Overall, all distillation objectives achieve comparable performance, with accuracy ranging from 85.92\% to 88.16\% and F1-scores from 0.8597 to 0.8822. Among them, Center loss attains the highest mean accuracy (88.16\%), F1-score (0.8822), precision (0.8894), and recall (0.8821). Compared with the student models distilled from PhoBERT-base, using the larger PhoBERT-large teacher does not consistently improve student performance, suggesting that increasing teacher capacity does not necessarily lead to more effective knowledge transfer for this task. The confusion matrices of the student models distilled from PhoBERT-large (Figure~\ref{fig13}) further illustrate that most disease classes maintain high prediction reliability while exhibiting confusion patterns similar to those observed for the PhoBERT-base teacher (see Figure \ref{figx13}.

\begin{table}
\centering
\scriptsize
\setlength{\tabcolsep}{3pt}
\caption{Performance of student models distilled from PhoBERT-large using different distillation objectives on the test set (the best indicators are in bold).}
\label{tab:dis_result_phobert_large}
\begin{tabular}{lcccc}
\hline
\multicolumn{1}{c}{\textbf{Metric}} & \multicolumn{1}{c}{\textbf{Acc (\%)}} & \multicolumn{1}{c}{\textbf{F1}} & \multicolumn{1}{c}{\textbf{Precision}} & \multicolumn{1}{c}{\textbf{Recall}} \\ \hline
\textbf{KL}                         & 88.06 ($\pm$0.96)                     & 0.8819 ($\pm$0.0099)            & \textbf{0.8941 ($\pm$0.0076)}                   & 0.8815 ($\pm$0.0096)                \\
\textbf{MSE}                        & 87.04 ($\pm$1.46)                     & 0.8708 ($\pm$0.0143)            & 0.8797 ($\pm$0.0152)                   & 0.8706 ($\pm$0.0139)                \\
\textbf{Cosine}                     & 87.76 ($\pm$1.27)                     & 0.8784 ($\pm$0.0115)            & 0.8911 ($\pm$0.0056)          & 0.8779 ($\pm$0.0127)                \\
\textbf{JSD}                        & 86.33 ($\pm$1.81)                     & 0.8646 ($\pm$0.0171)            & 0.8740 ($\pm$0.014)                    & 0.8636 ($\pm$0.0178)                \\
\textbf{SupCon}                     & 85.92 ($\pm$2.85)                     & 0.8597 ($\pm$0.0282)            & 0.8731 ($\pm$0.0263)                   & 0.8589 ($\pm$0.0285)                \\
\textbf{Triplet}                    & 86.53 ($\pm$1.31)                     & 0.8647 ($\pm$0.0138)            & 0.8765 ($\pm$0.01)                     & 0.8649 ($\pm$0.0138)   \\
\textbf{Center}                     & \textbf{88.16 ($\pm$3.33)}            & \textbf{0.8822 ($\pm$0.0343)}   & 0.8894 ($\pm$0.0358)                   & \textbf{0.8821 ($\pm$0.0337)}       \\
\hline
\end{tabular}
\end{table}

\subsubsection{Effectiveness of gated fusion}

Table~\ref{tab:student_without_gate_results} summarizes the performance of student models without the gated fusion module. Overall, classification performance remains comparable across different distillation objectives, with accuracy ranging from 85.10\% to 89.29\% and F1-scores from 0.8505 to 0.8929. Among the evaluated objectives, JSD attains the highest mean accuracy (89.29\%), F1-score (0.8929), precision (0.9060), and recall (0.8930), while Center also achieves competitive performance.

\begin{figure}
    \centering
    \includegraphics[width=1\linewidth]{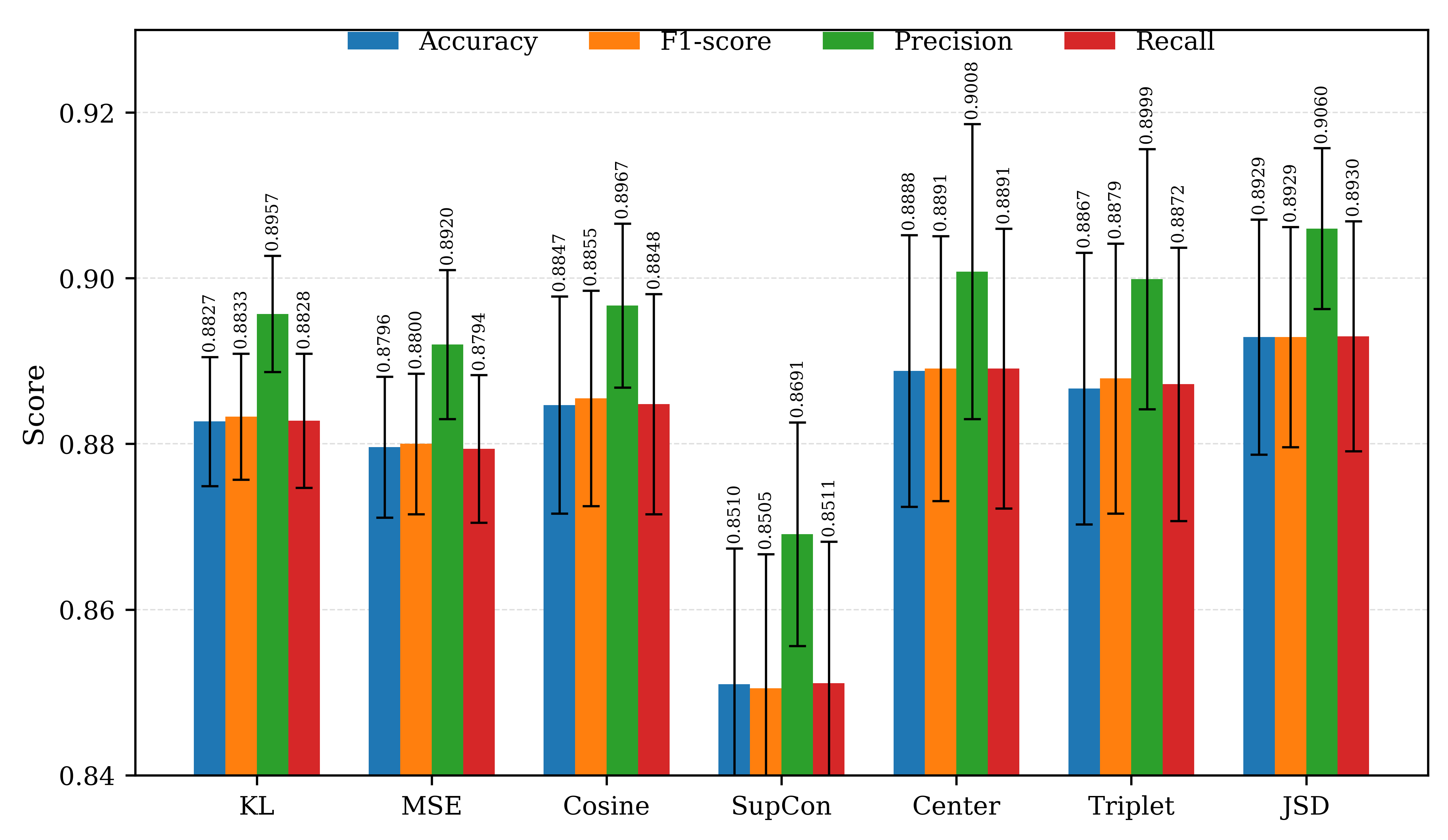}
    \caption{Performance of student models without gated fusion distilled from PhoBERT-base on the test set.}
    \label{figx14}
\end{figure}

Compared with the proposed student architecture equipped with gated fusion (Table~\ref{tab:student_loss_results}), removing the gated fusion module does not consistently reduce classification performance. Specifically, JSD and Center exhibit modest improvements after removing the gated fusion module, whereas KL, MSE, Cosine, and SupCon achieve comparable or slightly lower performance. Triplet also shows a moderate improvement in classification performance. These observations indicate that the contribution of the gated fusion module depends on the adopted distillation objective rather than providing a consistent performance gain across all objectives (see Figure \ref{figx14}).

\begin{table}[t]
\centering
\scriptsize
\setlength{\tabcolsep}{3pt}
\caption{Performance of student models without gated fusion distilled from PhoBERT-base on the test set (the best indicators are in bold).}
\label{tab:student_without_gate_results}
\begin{tabular}{lcccc}
\hline
\multicolumn{1}{c}{\textbf{Metric}} & \textbf{Acc (\%)}      & \textbf{F1}               & \textbf{Precision}        & \textbf{Recall}           \\
\hline
\textbf{KL}                         & 88.27 ($\pm$0.78)          & 0.8833 ($\pm$0.0076)          & 0.8957 ($\pm$0.0070)          & 0.8828 ($\pm$0.0081)          \\
\textbf{MSE}                        & 87.96 ($\pm$0.85)          & 0.8800 ($\pm$0.0085)          & 0.8920 ($\pm$0.0090)          & 0.8794 ($\pm$0.0089)          \\
\textbf{Cosine}                     & 88.47 ($\pm$1.31)          & 0.8855 ($\pm$0.0130)          & 0.8967 ($\pm$0.0099)          & 0.8848 ($\pm$0.0133)          \\

\textbf{SupCon}                     & 85.10 ($\pm$1.64)          & 0.8505 ($\pm$0.0162)          & 0.8691 ($\pm$0.0135)          & 0.8511 ($\pm$0.0171)          \\
\textbf{Center}                     & 88.88 ($\pm$1.64)          & 0.8891 ($\pm$0.0160)          & 0.9008 ($\pm$0.0178)          & 0.8891 ($\pm$0.0169)          \\
\textbf{Triplet}                    & 88.67 ($\pm$1.64)          & 0.8879 ($\pm$0.0163)          & 0.8999 ($\pm$0.0157)          & 0.8872 ($\pm$0.0165) \\

\textbf{JSD}                        & \textbf{89.29 ($\pm$1.42)} & \textbf{0.8929 ($\pm$0.0133)} & \textbf{0.9060 ($\pm$0.0097)} & \textbf{0.8930 ($\pm$0.0139)} \\
\hline
\end{tabular}
\end{table}

\begin{figure}
\centering
\includegraphics[width=\textwidth]{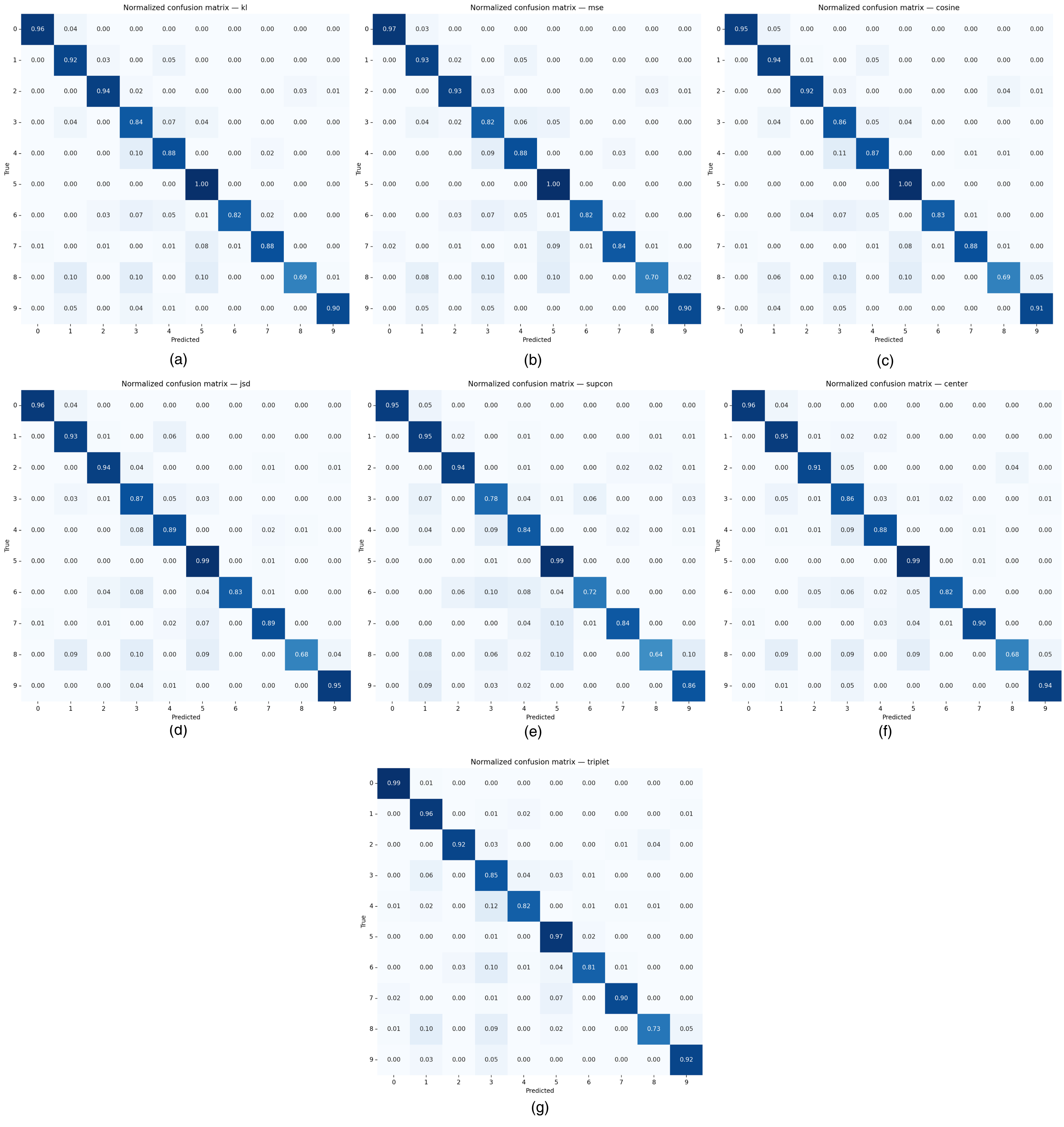}
\caption{Normalized confusion matrices for student models distilled from PhoBERT-base without the gated fusion module using different distillation objectives. Class indices: 0 = Acute Hepatopancreatic Necrosis Disease; 1 = White Feces Disease; 2 = White Spot Disease; 3 = Luminous Bacteria Disease; 4 = Loose Shell Syndrome; 5 = Black Gill Disease; 6 = Filamentous Bacterial Disease; 7 = Vitamin C Deficiency Disease; 8 = Yellow Head Disease; 9 = Taura Syndrome}
\label{fig14}
\end{figure}

Figure~\ref{fig14} presents the normalized confusion matrices of the student models distilled without the gated fusion module, revealing confusion patterns that remain largely consistent with those of the full SALT framework and strong diagonal dominance across most disease classes.
The conventional distillation objectives (KL, MSE, Cosine, and JSD) generally maintain stable class discrimination with strong diagonal dominance. Among the metric learning-based objectives, Center also produces a relatively clean confusion matrices with high correct classification rates across most classes. In contrast, SupCon exhibits the most dispersed off-diagonal errors, particularly for Filamentous Bacterial Disease (72\%) and Yellow Head Disease (64\%), indicating weaker class separability. JSD achieves the highest correct classification rate for Loose Shell Syndrome (89\%), whereas Center and Triplet both achieve the highest rate for Vitamin C Deficiency Disease (90\%). Triplet provides competitive class-wise performance but still exhibits moderate confusion among diseases with overlapping clinical symptom descriptions. Overall, removing the gated fusion module does not substantially alter the student model's decision patterns, suggesting that the gated fusion mechanism's contribution to class-level discrimination is objective-dependent rather than consistently beneficial across different distillation objectives.

\subsubsection{Effectiveness of linguistic augmentation}

\begin{figure}[H]
    \centering
    \includegraphics[width=0.8\linewidth]{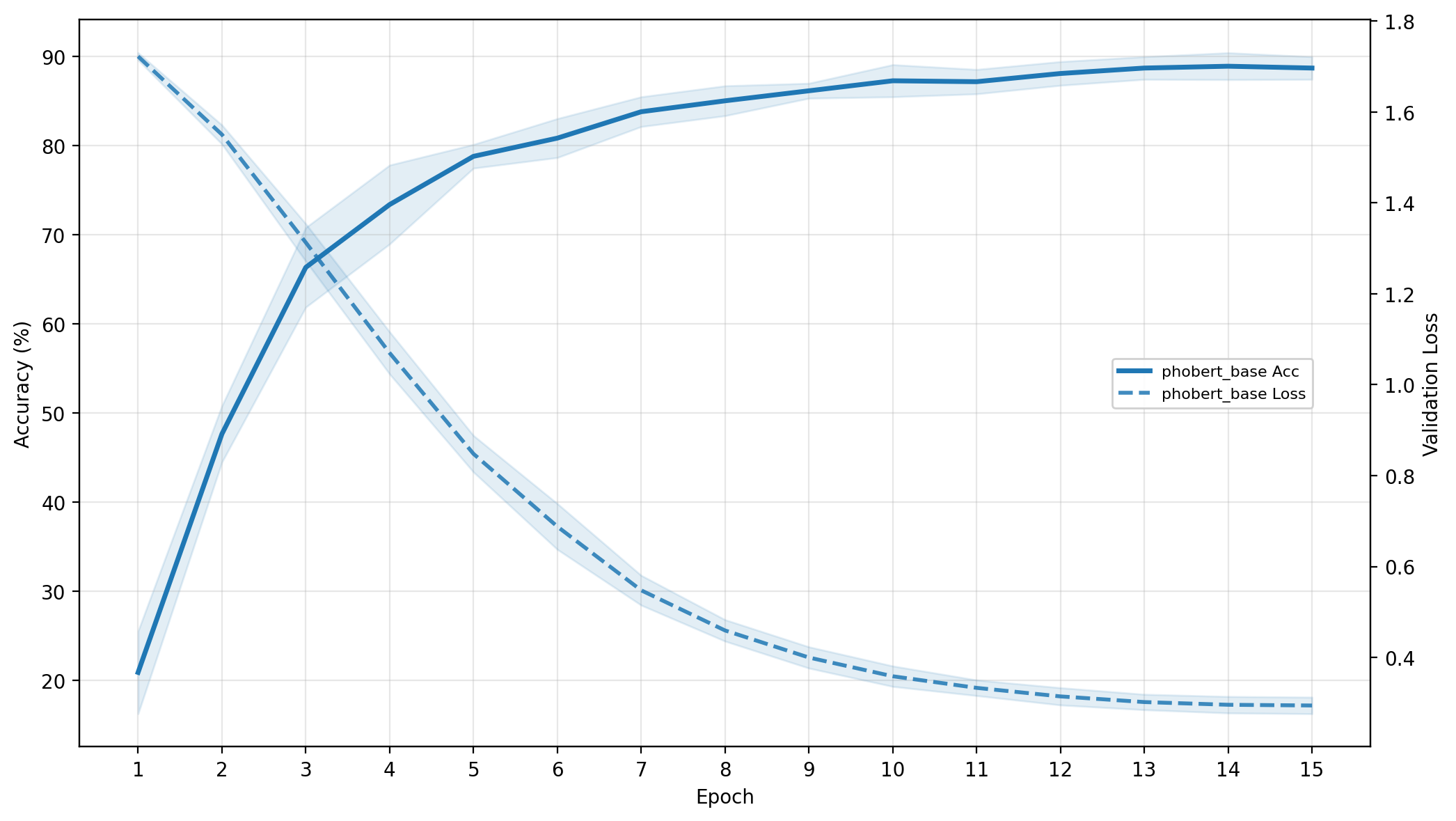}
    \caption{Validation accuracy and loss curves on the validation set for the PhoBERT-base teacher trained with linguistic augmentation.}
    \label{fig24}
\end{figure}

To investigate the effect of linguistic data augmentation, the PhoBERT-base teacher was fine-tuned using the augmented training set, achieving a validation accuracy of $87.96 \pm 0.72\%$ and a validation F1-score of $0.8771 \pm 0.0062$. Compared with the teacher trained on the original dataset, linguistic augmentation resulted in lower validation performance. The validation accuracy and loss curves over the training epochs, illustrated in Figure~\ref{fig24}, show that the model converges rapidly within the first few epochs, after which both the validation accuracy and loss remain relatively stable despite minor fluctuations in accuracy. These observations suggest that the performance degradation is unlikely to result from optimization instability and is more likely attributable to the characteristics of the augmented data.

\begin{figure}
    \centering
    \includegraphics[width=1\linewidth]{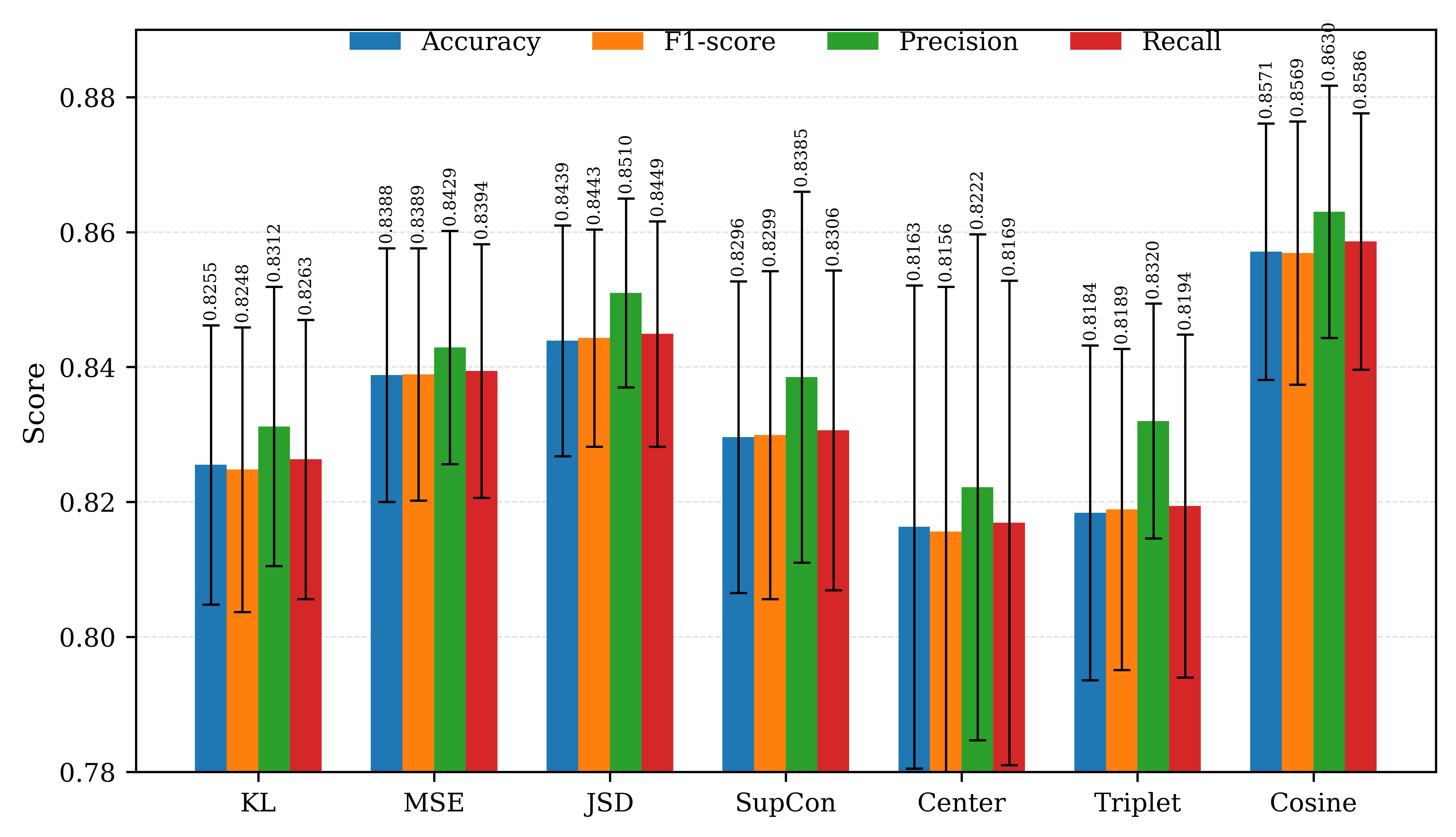}
    \caption{Performance of student models distilled from the linguistically augmented PhoBERT-base teacher on the test set.}
    \label{figx15}
\end{figure}

Table~\ref{tab:student_linguistic_results} summarizes the performance of the student models distilled from the linguistically augmented PhoBERT-base teacher. Among the evaluated objectives, Cosine achieves the best overall performance, obtaining the highest accuracy (85.71\%), F1-score (0.8569), precision (0.8630), and recall (0.8586), followed by JSD. The remaining objectives achieve comparable performance but at a lower level. Compared with the corresponding models trained on the original dataset (Table~\ref{tab:student_loss_results}), all distillation objectives consistently exhibit lower performance after linguistic augmentation. For example, the F1-score of the best-performing Cosine model decreases from 0.8935 to 0.8569, while the augmented teacher also underperforms its counterpart trained on the original dataset. These results suggest that the applied linguistic augmentation does not improve generalization for this task, likely because the generated samples are semantically similar to the original disease descriptions and provide limited additional discriminative information. Overall, the original manually curated dataset yields more effective supervision for both teacher training and subsequent knowledge distillation (see Figure \ref{figx15}).

\begin{table}[t]
\centering
\scriptsize
\setlength{\tabcolsep}{3pt}
\caption{Performance of student models distilled from the linguistically augmented PhoBERT-base teacher on the test set (the best indicators are in bold).}
\label{tab:student_linguistic_results}
\begin{tabular}{lcccc}
\hline
\textbf{Metric} & \textbf{Acc (\%)}      & \textbf{F1}               & \textbf{Precision}        & \textbf{Recall}           \\
\hline
\textbf{KL}                         & 82.55 (±2.07)          & 0.8248 (±0.0211)          & 0.8312 (±0.0207)          & 0.8263 (±0.0207)          \\
\textbf{MSE}                        & 83.88 (±1.88)          & 0.8389 (±0.0187)          & 0.8429 (±0.0173)          & 0.8394 (±0.0188)          \\
\textbf{JSD}                        & 84.39 (±1.71)          & 0.8443 (±0.0161)          & 0.8510 (±0.0140)          & 0.8449 (±0.0167)          \\
\textbf{SupCon}                     & 82.96 (±2.31)          & 0.8299 (±0.0243)          & 0.8385 (±0.0275)          & 0.8306 (±0.0237)          \\
\textbf{Center}                     & 81.63 (±3.58)          & 0.8156 (±0.0363)          & 0.8222 (±0.0375)          & 0.8169 (±0.0359)          \\
\textbf{Triplet}                    & 81.84 (±2.48)          & 0.8189 (±0.0238)          & 0.8320 (±0.0174)          & 0.8194 (±0.0254)  \\
\textbf{Cosine}                     & \textbf{85.71 (±1.90)} & \textbf{0.8569 (±0.0195)} & \textbf{0.8630 (±0.0187)} & \textbf{0.8586 (±0.0190)} \\
\hline
\end{tabular}
\end{table}

\begin{figure}
    \centering
    \includegraphics[width=1\linewidth]{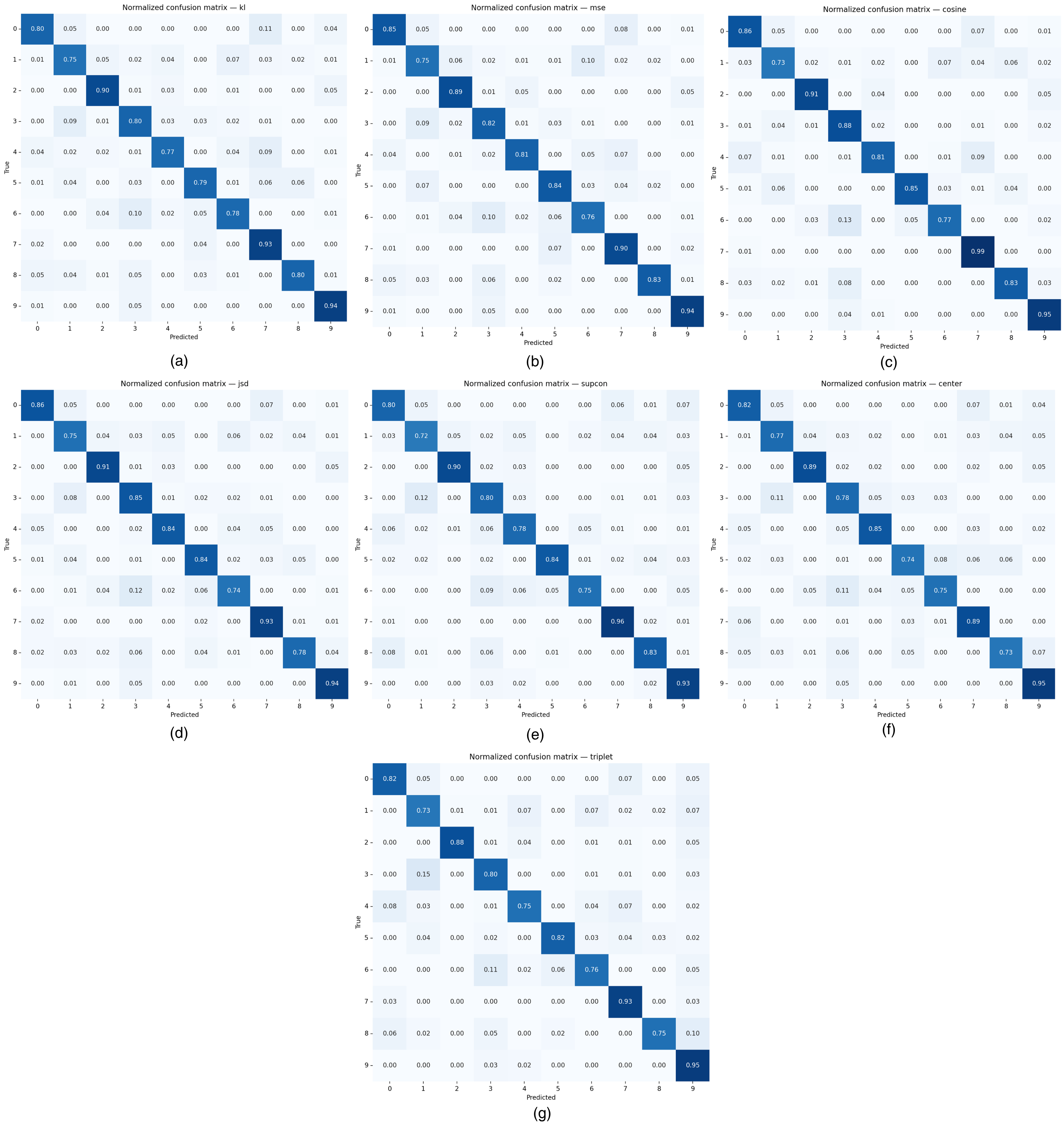}
    \caption{Normalized confusion matrices for student models distilled from the linguistically augmented PhoBERT-base teacher using different distillation objectives. Class indices: 0 = Acute Hepatopancreatic Necrosis Disease; 1 = White Feces Disease; 2 = White Spot Disease; 3 = Luminous Bacteria Disease; 4 = Loose Shell Syndrome; 5 = Black Gill Disease; 6 = Filamentous Bacterial Disease; 7 = Vitamin C Deficiency Disease; 8 = Yellow Head Disease; 9 = Taura Syndrome}
    \label{fig25}
\end{figure}

Figure~\ref{fig25} presents the normalized confusion matrices of the student models distilled from the linguistically augmented PhoBERT-base teacher using different distillation objectives. Overall, the confusion patterns remain broadly consistent across the evaluated objectives, with Cosine exhibiting the strongest class-level discrimination, followed by JSD, whereas SupCon produces the most dispersed off-diagonal errors. Compared with SupCon, Cosine achieves higher correct classification rates for several challenging disease classes, including Luminous Bacteria Disease (88\% vs. 80\%), Loose Shell Syndrome (81\% vs. 78\%), and Filamentous Bacterial Disease (77\% vs. 75\%), indicating better discrimination for diseases with overlapping clinical symptom descriptions. Relative to the corresponding student models distilled from the teacher trained on the original dataset, the models trained with linguistic augmentation exhibit more off-diagonal predictions and lower diagonal values across multiple disease classes, indicating reduced class separability. These observations are consistent with the quantitative results, suggesting that linguistic augmentation does not improve the effectiveness of knowledge distillation in the proposed framework.

\subsection{Explainability analysis}
\subsubsection{Qualitative explainability}
\begin{figure}[H]
    \centering
    \includegraphics[width=0.8\linewidth]{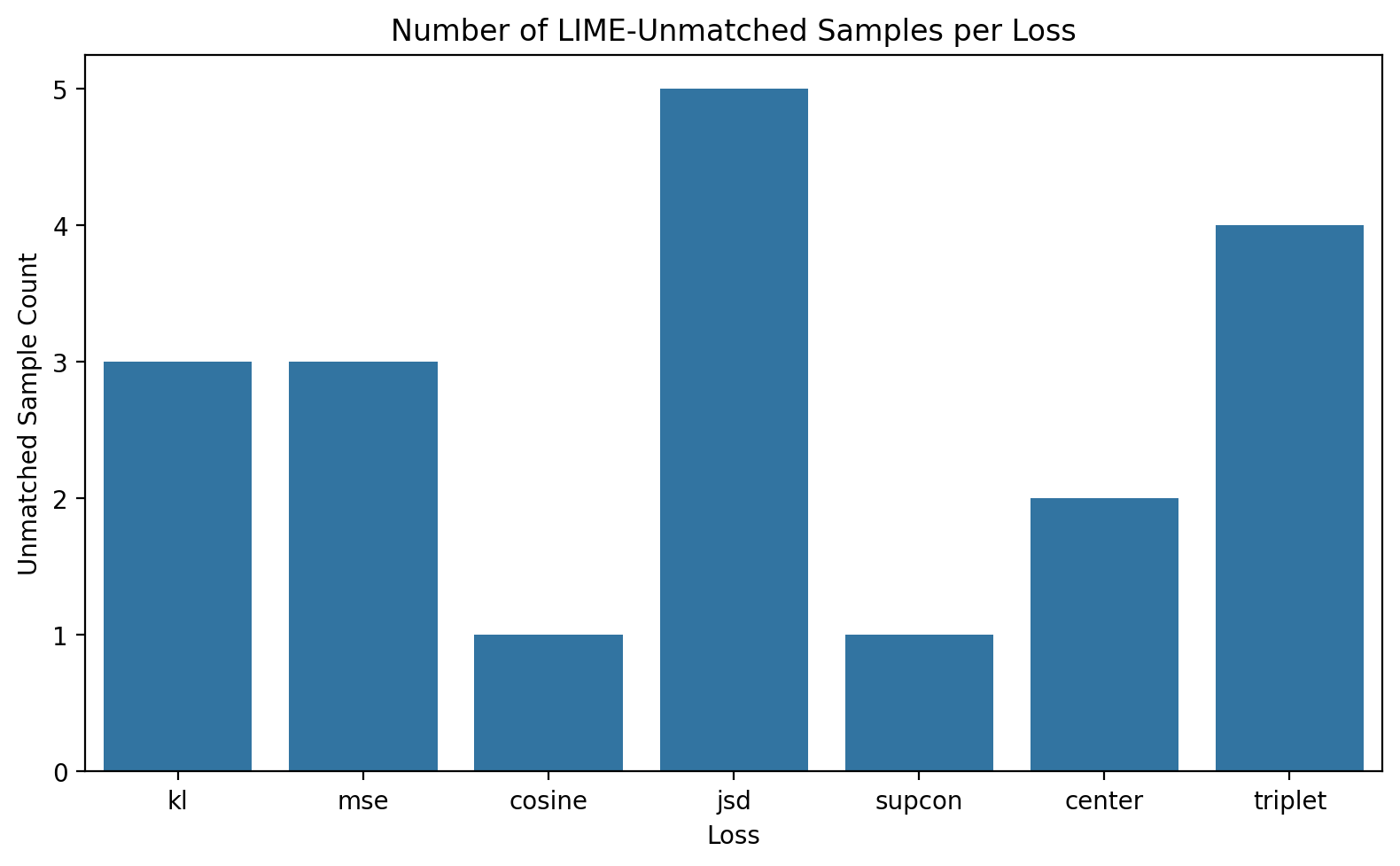}
    \caption{Number of LIME-Unmatched samples for each distillation objective}
    \label{fig18}
\end{figure}

Figure~\ref{fig18} summarizes the number of LIME-unmatched samples for each distillation objective, where a LIME-unmatched sample is defined as a test instance for which the Top-5 most important words identified by LIME for the student model do not match those identified for the corresponding teacher prediction. The number of unmatched explanations is low across all loss functions, indicating that the distilled student models generally produce explanation patterns similar to those of the teacher model. Among the evaluated objectives, JSD yields the largest number of unmatched samples (5), followed by Triplet (4). KL and MSE each produce three unmatched samples, Center produces two, whereas Cosine and SupCon each result in only one unmatched sample, indicating the highest explanation consistency with the teacher among the evaluated distillation objectives.

\begin{figure}[H]
    \centering
    \includegraphics[width=0.9\linewidth]{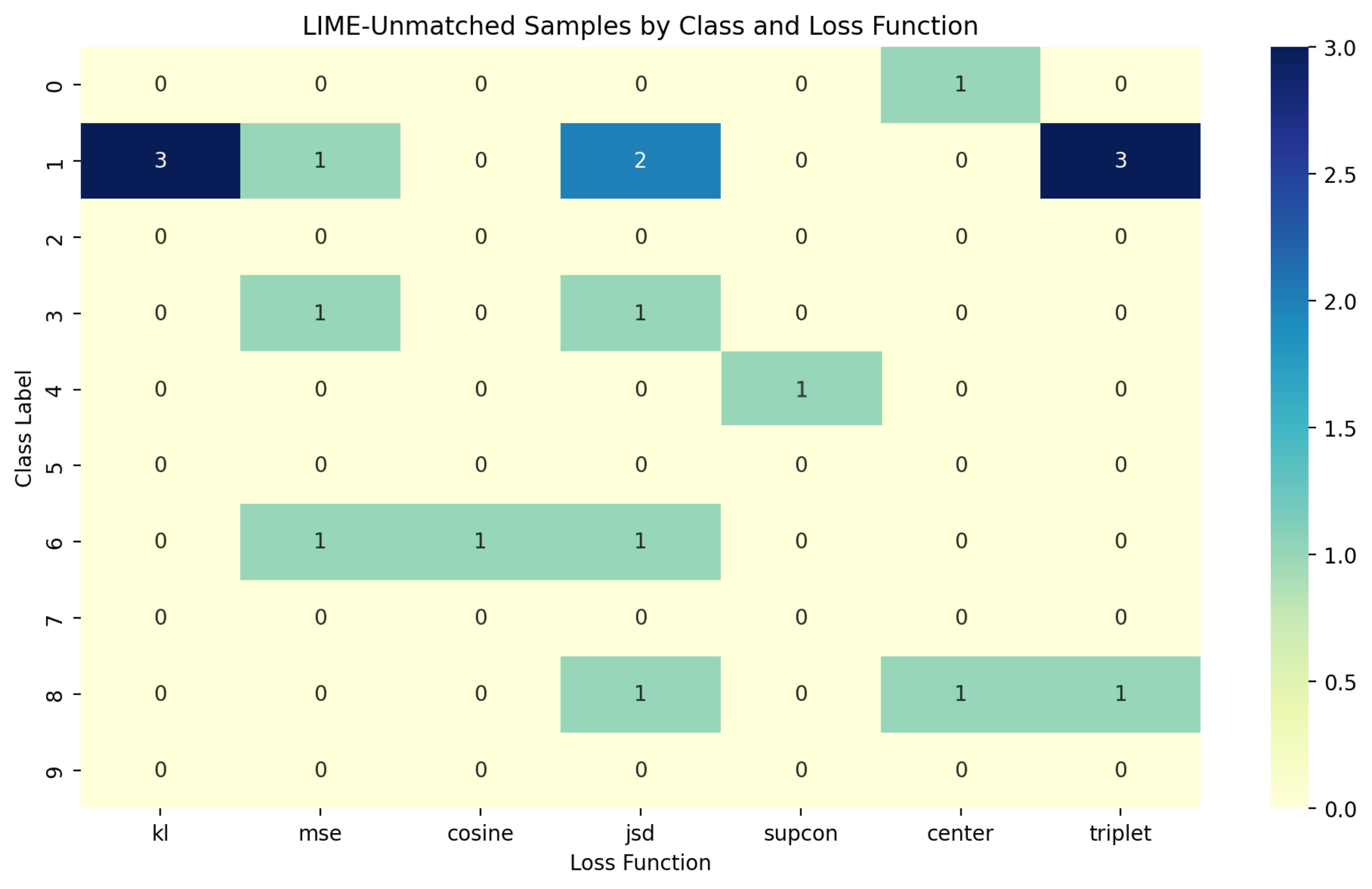}
    \caption{Heatmap of LIME-Unmatched samples by class and distillation objective. Class indices: 0 = Acute Hepatopancreatic Necrosis Disease; 1 = White Feces Disease; 2 = White Spot Disease; 3 = Luminous Bacteria Disease; 4 = Loose Shell Syndrome; 5 = Black Gill Disease; 6 = Filamentous Bacterial Disease; 7 = Vitamin C Deficiency Disease; 8 = Yellow Head Disease; 9 = Taura Syndrome}
    \label{fig19}
\end{figure}

Figure~\ref{fig19} presents the distribution of LIME-unmatched samples across disease classes. Most unmatched cases are concentrated in White Feces Disease, where KL and Triplet each produce three unmatched samples, while JSD produces two. Additional unmatched samples are sparsely distributed across Luminous Bacteria Disease, Filamentous Bacterial Disease, Yellow Head Disease, and Acute Hepatopancreatic Necrosis Disease. In contrast, Cosine and SupCon each exhibit only a single unmatched sample, suggesting more consistent explanation behavior across disease categories. Overall, the unmatched explanations are limited to only a few disease classes, indicating that the explanation patterns learned by the distilled students remain largely aligned with those of the teacher model.

\begin{figure}
    \centering
    \includegraphics[width=0.95\linewidth]{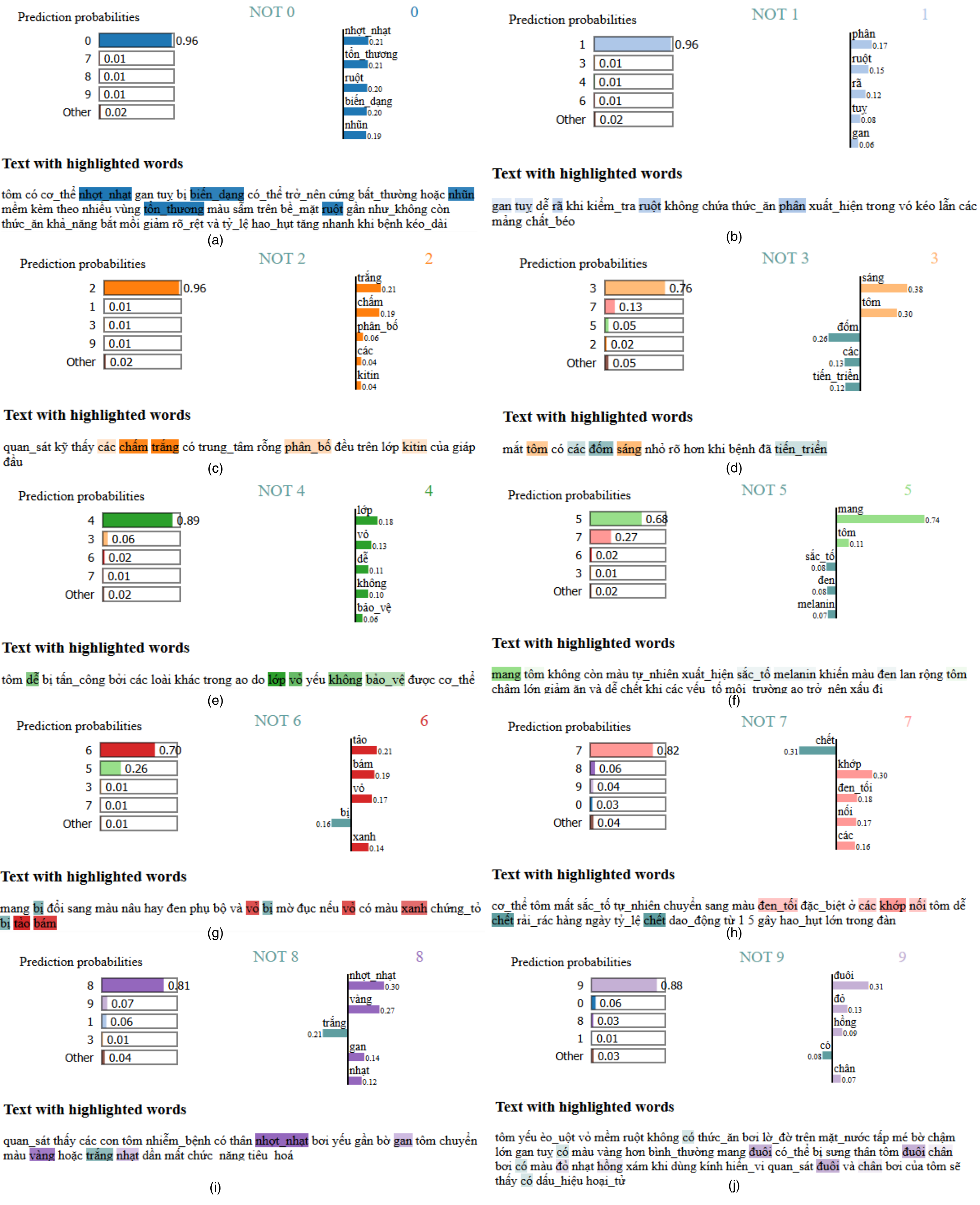}.
    \caption{Representative LIME visualizations for the best-performing student model (Cosine KD). One correctly classified example is shown for each disease class: (a) Acute Hepatopancreatic Necrosis Disease; (b) White Feces Disease; (c) White Spot Disease; (d) Luminous Bacteria Disease; (e) Loose Shell Syndrome; (f) Black Gill Disease; (g) Filamentous Bacterial Disease; (h) Vitamin C Deficiency Disease; (i) Yellow Head Disease; and (j) Taura Syndrome}
    \label{fig20}
\end{figure}

For qualitative evaluation, Figure~\ref{fig20} presents LIME explanations for one representative sample from each disease class using the best-performing student model trained with Cosine knowledge distillation. Across all ten classes, the highlighted words closely correspond to clinically meaningful symptoms and disease-specific terminology. For example, Acute Hepatopancreatic Necrosis Disease emphasizes keywords related to hepatopancreatic abnormalities, such as ``nhợt nhạt'' (pale), ``tổn thương'' (damaged), ``ruột'' (gut), and ``biến dạng'' (deformation). White Feces Disease is characterized by terms including ``phân'' (feces), ``ruột'' (gut), and ``tụy'' (hepatopancreas), whereas White Spot Disease focuses on the highly discriminative keywords ``chấm'' (spots), ``trắng'' (white), and ``phân bố'' (distributed). Similarly, the explanations for the remaining classes consistently highlight representative symptoms, including the keywords ``đốm'' (spots) and ``sáng'' (luminous) for Luminous Bacteria Disease, ``vỏ'' (shell) for Loose Shell Syndrome, ``mang'' (gills) for Black Gill Disease, ``tảo'' (algal) for Filamentous Bacterial Disease, ``đen tối'' (dark coloration) for Vitamin C Deficiency Disease, ``vàng'' (yellow) and ``nhợt nhạt'' (pale) for Yellow Head Disease, and ``đuôi'' (tail) together with ``đỏ'' (red) for Taura Syndrome. These visualizations suggest that the distilled student bases its predictions on clinically meaningful disease-specific features.

\subsubsection{Global explanation using SHAP}

To analyze global feature attribution, we computed the mean absolute SHAP value of each student model. Table~\ref{tab:tableVIII} summarizes the average SHAP score, SHAP standard deviation, and F1-score for the evaluated distillation objectives. JSD exhibits the highest average SHAP value (0.0212) together with the largest standard deviation (0.0457), indicating stronger overall feature attribution and greater variability in feature contributions across samples. In contrast, Cosine and Triplet produce the lowest average SHAP values (0.0097 and 0.0093, respectively), with Triplet also showing the smallest standard deviation, suggesting more concentrated attribution patterns across samples.

\begin{table}[H]
\centering
\footnotesize
\caption{Comparison of average SHAP score (Avg SHAP), SHAP standard deviation (SHAP Std.), and F1-Score across different distillation objectives}
\label{tab:tableVIII}
\begin{tabular}{lccccccc}
\hline
\textbf{Metric} & \textbf{KL} & \textbf{MSE} & \textbf{Cosine} & \textbf{JSD} & \textbf{SupCon} & \textbf{Center} & \textbf{Triplet} \\
\hline
Avg SHAP
& 0.0179
& 0.0186
& 0.0097
& \textbf{0.0212}
& 0.0131
& 0.0168
& 0.0093 \\

SHAP Std.
& 0.0396
& 0.0399
& 0.0204
& 0.0457
& 0.0283
& 0.0388
& 0.0174 \\

F1-score
& 0.8880
& 0.8822
& \textbf{0.8935}
& 0.8773
& 0.8601
& 0.8743
& 0.8737 \\
\hline
\end{tabular}
\end{table}

\begin{figure}
    \centering
    \includegraphics[width=0.8\linewidth]{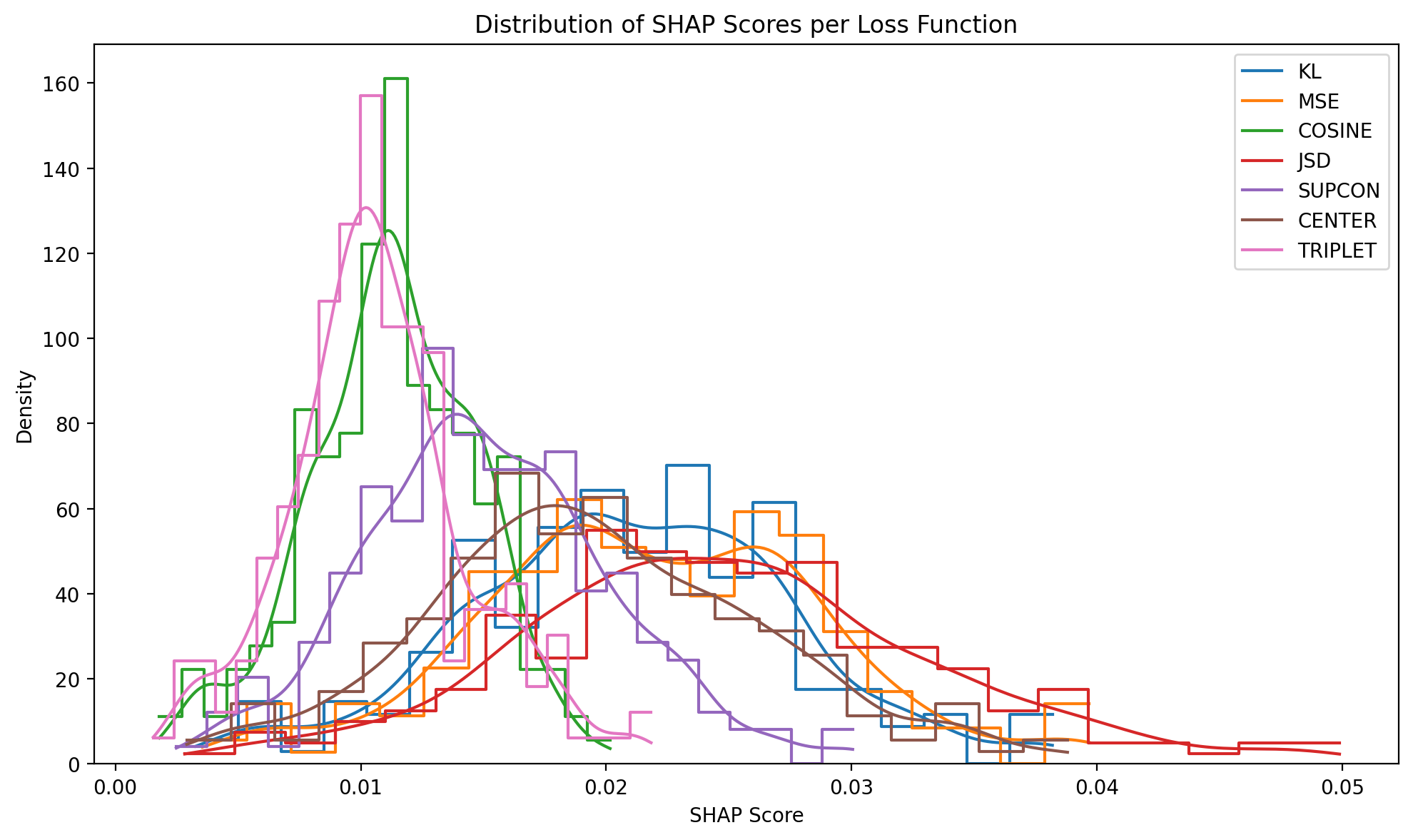}
    \caption{Distribution of SHAP scores per loss function.}
    \label{fig21}
\end{figure}

Figure~\ref{fig21} illustrates the distribution of SHAP values for the evaluated distillation objectives. JSD exhibits the broadest distribution with a pronounced right tail, indicating a wider range of feature attribution magnitudes across samples. In contrast, Cosine and Triplet display narrower distributions centered at lower SHAP values, reflecting more concentrated attribution patterns. KL, MSE, and Center exhibit intermediate distributions, whereas SupCon shows a moderately concentrated distribution with relatively low SHAP values.

\begin{figure}
    \centering
    \includegraphics[width=0.8\linewidth]{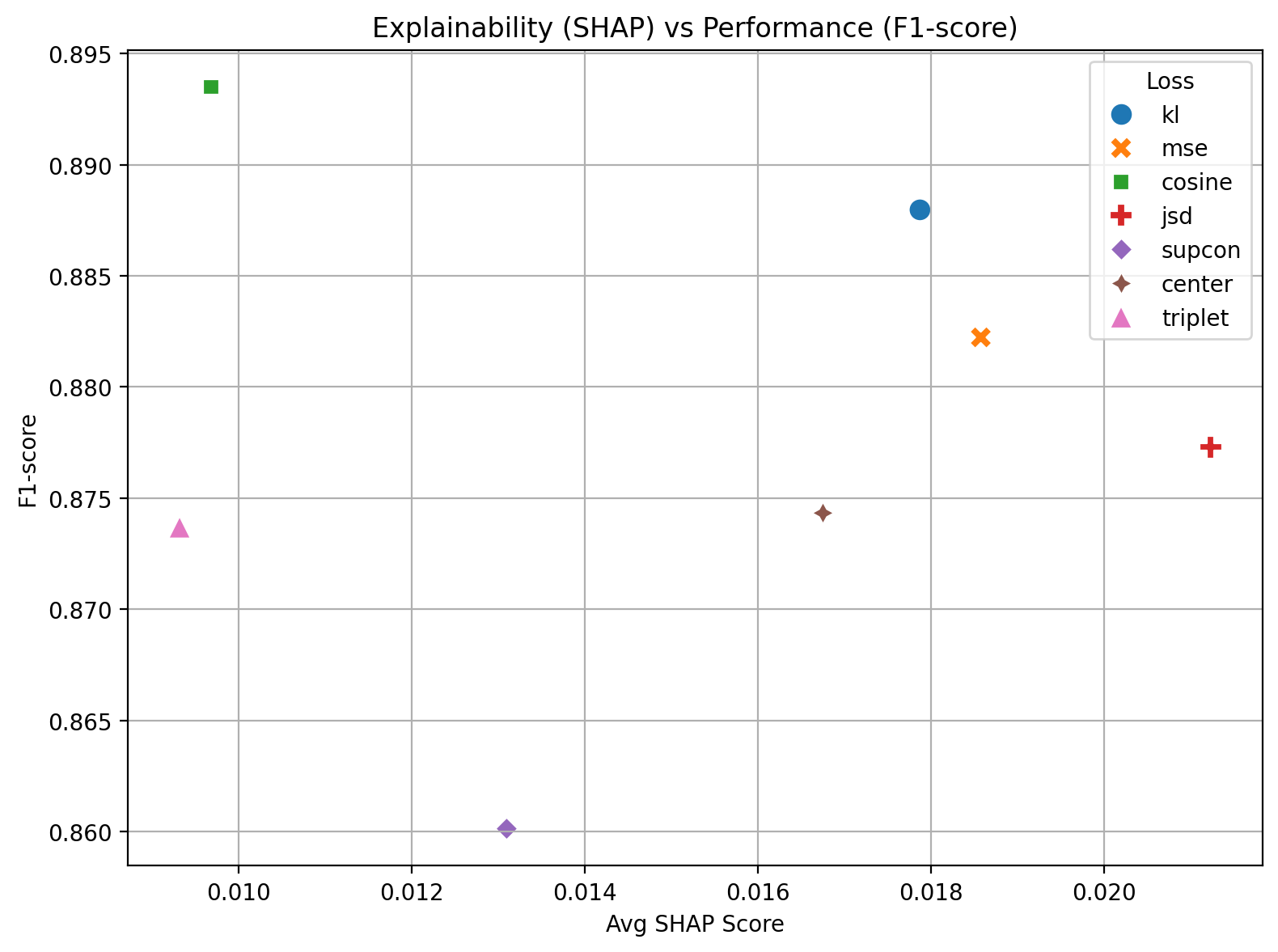}
    \caption{Explainability vs. Performance}
    \label{fig22}
\end{figure}

The relationship between global feature attribution and classification performance is shown in Figure~\ref{fig22}. The results indicate that larger average SHAP values do not necessarily correspond to better classification performance. Although JSD achieves the highest average SHAP score, its F1-score is lower than those of KL, MSE, and Cosine. Conversely, Cosine attains the highest F1-score (0.8935) despite having one of the lowest average SHAP values. SupCon records both relatively low feature attribution and the lowest F1-score among the evaluated methods. These observations suggest that global feature attribution and predictive performance capture complementary aspects of model behavior rather than exhibiting a simple linear relationship.

Combining the quantitative performance, teacher--student explanation consistency, and global feature attribution analyses, the results indicate that different distillation objectives exhibit different strengths. Cosine loss achieves the highest classification performance while maintaining high explanation consistency with the teacher model. JSD produces the strongest global feature attribution according to the average SHAP value, but does not achieve the best classification performance. Taken together, the results indicate that global feature attribution and predictive performance capture complementary aspects of model behavior and should be considered jointly when evaluating knowledge distillation methods for shrimp disease text classification.

\subsubsection{Keyword deletion faithfulness}

The keyword deletion results support the faithfulness of the generated explanations. The clear separation between explanation-guided deletion and random deletion indicates that the keywords identified by LIME are important for the model predictions.

\begin{figure}
    \centering
    \includegraphics[width=\linewidth]{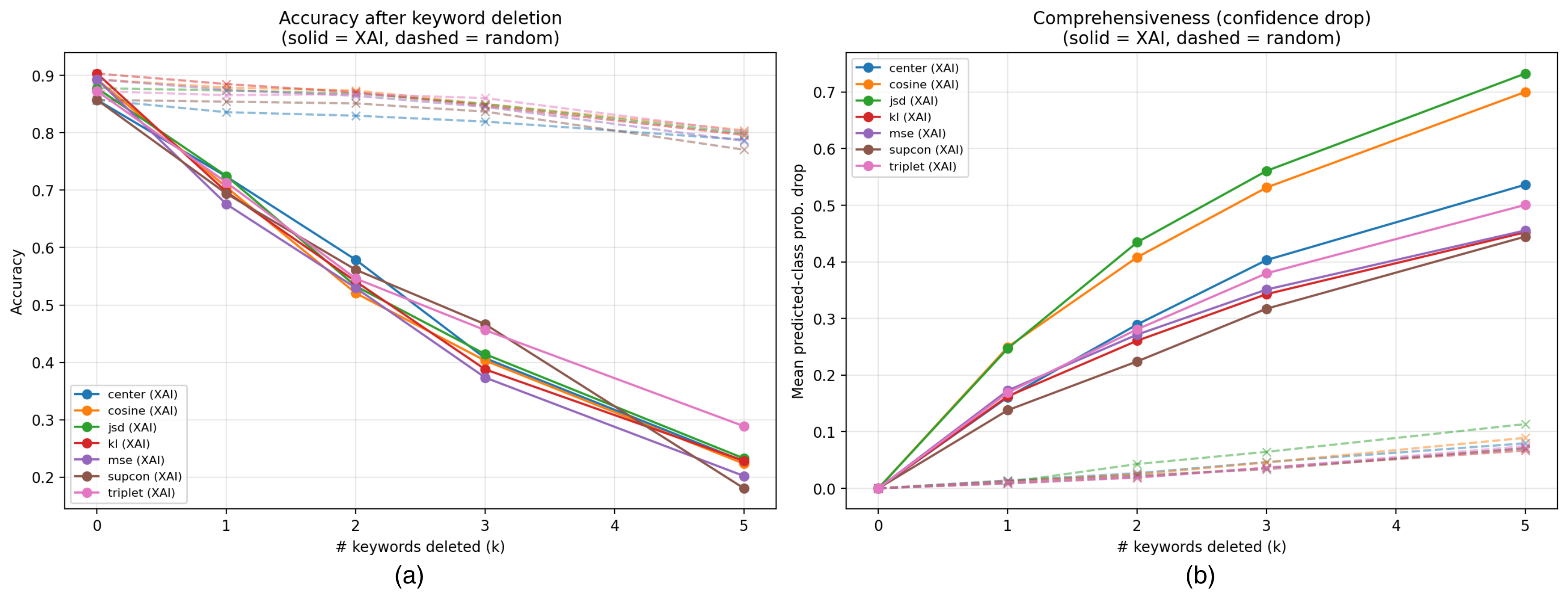}
    \caption{Keyword deletion faithfulness analysis: (a) Classification accuracy after removing the Top-$k$ keywords identified by LIME (Solid Lines) and randomly selected keywords (Dashed Lines), (b) Comprehensiveness is measured as the average reduction in the predicted-class probability after keyword removal.}
    \label{fig23}
\end{figure}

Figure~\ref{fig23}(a) presents the results of the keyword deletion test. Removing the Top-$k$ keywords identified by LIME (solid lines) consistently causes a substantially larger reduction in classification accuracy than deleting the same number of randomly selected words (dashed lines) across all distillation objectives. The performance gap widens as more keywords are removed, indicating that the features highlighted by LIME play an important role in the model predictions.

The comprehensiveness analysis in Figure~\ref{fig23}(b) further supports this observation. Removing the explanation-guided keywords produces substantially larger reductions in the predicted-class confidence than random deletion. JSD exhibits the largest confidence drop, followed by Cosine, whereas SupCon and KL show comparatively smaller reductions, indicating differences in the importance of the identified features across distillation objectives.

\subsection{Computational efficiency}
Table~\ref{tab:efficiency} summarizes the computational efficiency of the PhoBERT-base teacher, the best-performing distilled student (Cosine KD), and the baseline models. Model size corresponds to the memory footprint after loading the model into CPU memory.

\begin{table}[t]
\centering
\caption{Computational efficiency comparison of the Teacher, Student, and baseline models. $^\dagger$GFLOPs correspond only to the DNN classifier; sparse TF-IDF vectorization is excluded.}
\label{tab:efficiency}
\resizebox{\textwidth}{!}{
\begin{tabular}{lcccccc}
\hline
\textbf{Model} &
\textbf{Params (M)} &
\textbf{Size (MB)} &
\textbf{GFLOPs} &
\textbf{GPU Latency (ms)} &
\textbf{CPU Latency (ms)} &
\textbf{Throughput (samples/s)} \\
\hline
PhoBERT-base (Teacher)
& 135.01
& 515.01
& 43.49
& $13.34 \pm 0.64$
& $518.01 \pm 6.43$
& 1.93 \\

Student (Cosine KD)
& 79.88
& 304.73
& 14.50
& $3.83 \pm 0.15$
& $175.67 \pm 1.12$
& 5.69 \\

SFT
& 79.88
& 304.73
& 14.50
& $3.91 \pm 0.04$
& $175.67 \pm 1.12$
& 5.69 \\

TF-IDF + DNN
& 11.45
& 131.05
& -$^\dagger$
& $0.46 \pm 0.44$
& $1.00 \pm 0.15$
& 996.21 \\

F2LLM-v2-80M
& 80.38
& 306.64
& 32.21
& $5.53 \pm 0.13$
& $67.41 \pm 0.11$
& 14.83 \\
\hline
\end{tabular}
}
\vspace{1mm}

\end{table}

The distilled student substantially reduces computational cost while maintaining competitive predictive performance. Specifically, the computational requirement is reduced from 43.49 to 14.50 GFLOPs (66.7\%), while GPU inference latency decreases from 13.34 ms to 3.83 ms per sample. CPU latency is similarly reduced from 518.01 ms to 175.67 ms, increasing throughput from 1.93 to 5.69 samples/s. In addition, the model size decreases from 515.01 MB to 304.73 MB.

The distilled student and the supervised fine-tuning (SFT) baseline share the same architecture and therefore exhibit nearly identical computational characteristics, including parameter count, model size, GFLOPs, latency, and throughput. The performance improvement of the distilled student is therefore attributable to knowledge distillation rather than increased model capacity or computational resources.

Among the external baselines, TF--IDF + DNN achieves the lowest computational cost, with only 11.45M parameters, a model size of 131.05 MB, negligible GFLOPs for the classifier, and the highest throughput (996.21 samples/s). However, this computational efficiency comes at the expense of lower predictive performance. In contrast, F2LLM-v2-80M has a parameter count and model size comparable to those of the distilled student but requires substantially higher computational complexity (32.21 vs.\ 14.50 GFLOPs) and longer GPU inference latency (5.53 vs.\ 3.83 ms), while providing inferior classification performance.

Overall, the proposed student provides a favorable balance between predictive performance and computational efficiency among the evaluated transformer-based models. Although these measurements demonstrate consistent efficiency gains on desktop hardware, evaluating deployment on mobile devices and other resource-constrained platforms remains an important direction for future work.

\section{Discussion}

The experimental results demonstrate that knowledge distillation provides consistent benefits for Vietnamese shrimp disease text classification. Although the distilled student contains only 79.88M parameters, it achieves classification performance comparable to that of the PhoBERT-base teacher while substantially reducing computational cost. Among the evaluated teacher models, PhoBERT-base provides a more suitable source of knowledge than mBART by achieving higher predictive performance, together with considerably lower inference latency. These findings indicate that an effective teacher should provide not only strong predictive capability but also transferable semantic representations that can be efficiently learned by a compact student model.

Among the evaluated distillation objectives, the conventional distribution-based losses consistently outperform the metric learning-based objectives. In particular, Cosine loss achieves the best overall classification performance while maintaining the highest level of explanation consistency with the teacher, according to the LIME analysis. Preserving the angular similarity between teacher and student representations, therefore, appears to be more effective than enforcing explicit metric constraints for this text classification task. In contrast, SupCon, Center, and Triplet generally produce lower predictive performance, suggesting that metric-learning objectives are less suitable for transferring semantic knowledge from the teacher under the proposed framework.

The comparative experiments further provide several insights into the knowledge distillation process. First, replacing PhoBERT-base with the larger PhoBERT-large teacher does not consistently improve student performance, despite the larger teacher's higher standalone performance. This observation suggests that increasing teacher capacity does not necessarily improve knowledge transfer, since more complex representation spaces may be more difficult for a lightweight student to approximate. Second, removing the gated fusion module does not consistently degrade performance across distillation objectives, indicating that its contribution depends on the selected distillation loss rather than providing a universal performance gain. Third, linguistic augmentation consistently reduces the performance of both the teacher and distilled students. The generated descriptions are likely to remain semantically close to the original samples, providing limited additional discriminative information while introducing redundant training examples.

The baseline comparisons further highlight the effectiveness of the proposed framework. Supervised fine-tuning consistently underperforms the distilled student despite using the same architecture, demonstrating that the observed performance gains originate from knowledge transfer rather than increased model capacity. The TF--IDF + DNN baseline achieves competitive classification performance but relies on sparse lexical representations that struggle to distinguish diseases with overlapping clinical symptom descriptions and requires substantially longer inference time. Meanwhile, F2LLM-v2-80M yields lower predictive performance despite requiring substantially more floating-point operations and higher GPU inference latency than the distilled student.

The explainability analyses provide complementary evidence supporting the proposed framework. LIME demonstrates that the distilled students generally preserve explanation patterns similar to those of the teacher. Cosine exhibits high teacher--student explanation consistency while also achieving the best overall classification performance. The qualitative LIME visualizations further show that the highlighted keywords correspond closely to clinically meaningful disease symptoms. Global SHAP analysis reveals that different distillation objectives learn distinct feature attribution patterns. In particular, JSD achieves the highest average SHAP score, whereas Cosine achieves the best classification performance. These results indicate that stronger global feature attribution alone does not necessarily imply superior predictive performance, and that explainability should be evaluated alongside predictive performance. Finally, the keyword deletion experiments provide additional evidence for the faithfulness of the generated explanations, as removing the LIME-identified keywords consistently causes much larger performance degradation than random deletion.

Despite these promising results, several limitations remain. First, the ShrimpCap dataset is relatively small and may not fully capture the diversity of disease stages, farming conditions, and regional linguistic variations encountered in real-world shrimp farming. Second, all experiments were conducted on desktop hardware, and the deployment efficiency of the proposed student model on mobile or embedded devices has not yet been evaluated. Finally, the study focuses exclusively on Vietnamese textual symptom descriptions. Extending the framework to multilingual datasets and multimodal disease diagnosis that combines textual descriptions with shrimp images represents an important direction for future research.

\section{Conclusion}
This study proposed SALT, an explainable multi-loss knowledge distillation framework for Vietnamese shrimp disease classification based on textual symptom descriptions. In addition, ShrimpCap was constructed as the first Vietnamese dataset dedicated to shrimp disease symptom descriptions, comprising 979 manually validated samples covering ten major shrimp diseases. Experimental results demonstrate that PhoBERT-base provides a more effective teacher than mBART, achieving superior predictive performance while enabling more effective knowledge transfer to the compact student model. Among the seven evaluated distillation objectives, Cosine loss achieves the best overall classification performance while maintaining high teacher--student explanation consistency. Comparative experiments further show that increasing teacher capacity with PhoBERT-large does not consistently improve the distilled student and that conventional linguistic augmentation provides limited benefit for this task. The proposed framework also consistently outperforms supervised fine-tuning and competitive baseline models while substantially reducing computational cost relative to the teacher model. Besides that, the explainability analyses provide complementary insights into the behavior of the distilled models. LIME demonstrates that the student generally preserves explanation patterns similar to those of the teacher, while the qualitative visualizations highlight clinically meaningful disease-specific keywords. SHAP analysis reveals that different distillation objectives learn distinct global feature attribution patterns, with JSD achieving the highest average SHAP score despite not producing the best classification performance. Furthermore, the keyword deletion experiments provide additional evidence for the faithfulness of the generated explanations. These findings indicate that predictive performance, explanation consistency, global feature attribution, and faithfulness should be considered jointly when evaluating explainable knowledge distillation models.

\section{Future work}
Based on the experimental results and existing limitations, several further research directions are proposed to expand the model's applicability and enhance its effectiveness. Firstly, it involves expanding and diversifying the data set by collecting data from various farming areas, at different disease stages, and with more diverse expression styles. Secondly, the deployment and evaluation of the student model in real-world environments are necessary to confirm its advantages in terms of speed, memory consumption, and energy efficiency under limited-resource conditions. Thirdly, other advanced distillation techniques, such as semi-supervised and unsupervised knowledge distillation, will be considered to reduce the need for labeled data, thereby saving costs and implementation time. Finally, research on intrinsic explainability should also be conducted to reduce dependence on post-hoc methods, such as LIME or SHAP, while enhancing the model's transparency and reliability.




\section*{Acknowledgements}
Khang Nguyen Quoc was supported by the Hyundai Motor Chung Mong-Koo Foundation Global Scholarship (GSS-25-02120).

\section*{Data availability}
\sloppy The dataset used in this study is publicly available at \url{https://huggingface.co/datasets/nqanh312/ShrimpCap}


\section*{Declarations}
\textbf{Anh Nguyen Quynh} contributed to conceptualization, data curation, methodology, software, visualization, and writing – original draft. \textbf{Khang Nguyen Quoc} contributed to methodology, supervision, validation, and writing – review \& editing. \textbf{Luyl-Da Quach} contributed to project administration, supervision, validation, and writing – review \& editing. 

\bibliographystyle{elsarticle-num-names} 
\bibliography{cas-refs}






\end{document}